\documentclass{article}

\usepackage[preprint]{neurips_2026}
\usepackage{amsmath}
\usepackage{multirow}
\usepackage{graphicx}
\usepackage{xcolor}
\usepackage{comment}
\usepackage{amssymb}
\usepackage{subcaption}
\usepackage{booktabs}
\usepackage{tabularx}
\usepackage{subcaption}
\usepackage{pgfplots}
\pgfplotsset{compat=1.18}
\usepgfplotslibrary{groupplots}
\usepackage[normalem]{ulem}

\usepackage[utf8]{inputenc} 
\usepackage[T1]{fontenc}    
\usepackage{hyperref}       
\usepackage{url}            
\usepackage{booktabs}       
\usepackage{amsfonts}       
\usepackage{nicefrac}       
\usepackage{microtype}      
\usepackage[table]{xcolor}
\usepackage{wrapfig}
\usepackage{xcolor}
\usepackage{pifont}
\usepackage{tikz}
\usepackage{pgfplots}
\pgfplotsset{width=10cm,compat=1.18}

\definecolor{babyblue}{RGB}{204, 229, 255}
\definecolor{babyblue2}{RGB}{124, 239, 255}

\newcommand{\cmark}{\textcolor{green!60!black}{\ding{51}}} 
\newcommand{\xmark}{\textcolor{red!70!black}{\ding{55}}}   

\title{\textsc{hidra}: Hierarchical Dual-Routing Attention for  Replay-Free Lifelong Imitation Learning}

\author{%
Fanqi Yu$^{1,3}$ \quad
Matteo Tiezzi$^{4}$ \quad
Cigdem Beyan$^{1,2}$ \quad
Tommaso Apicella$^{4}$ \quad
Vittorio Murino$^{1,2}$ \\[0.5em]
$^{1}$AI for Good (AIGO) Research Unit, Istituto Italiano di Tecnologia, Genoa, Italy \\
$^{2}$Department of Computer Science, University of Verona, Verona, Italy \\
$^{3}$DITEN, University of Genoa, Genoa, Italy \\
$^{4}$Pattern Analysis and Computer Vision (PAVIS), Istituto Italiano di Tecnologia, Genoa, Italy
}
  
\begin{document}

\maketitle
\begin{center}
{\small \textcolor{red}{This is the authors' preprint, accepted to the NeurIPS 2026 Conference. \\ The final version is available in the NeurIPS 2026 proceedings.}}
\end{center}

\begin{abstract}
Robotic agents operating in real-world environments must adapt continuously from a stream of multimodal demonstrations, often under constraints that preclude storing or revisiting past data. In this replay-free, single-pass lifelong imitation learning setting, sequential updates can progressively distort the representation space, making expert-based approaches particularly sensitive to routing interference, causing expert selection to degrade over time. 
We propose HIDRA, a structured routing framework that mitigates this issue through a hierarchical dual-attention mechanism, which decouples instruction-level expert retrieval from context-dependent refinement. To further stabilize routing under representation drift, HIDRA introduces a key-level geometric regularization that enforces both alignment within tasks and separation across tasks. The resulting approach enables reliable expert selection while supporting an expandable set of experts for continual adaptation.
We evaluate HIDRA on several LIBERO benchmark suites, including more challenging variants with heterogeneous task sequences and paraphrased language instructions. Our approach consistently outperforms both replay-free and replay-based baselines, improving AUC and forward transfer while maintaining low forgetting, with stronger gains under high interference and language variation. The code is available at: \url{https://github.com/yfqi/HIDRA}.

\end{abstract}

\section{Introduction}
\label{sec:intro}

Robotic agents deployed in the real world must adapt continuously as they interact with dynamic, non-stationary environments. Unlike offline imitation learning pipelines that rely on curated datasets and multi-epoch optimization, real robots receive an unbounded stream of demonstrations and observations, often under constraints that prevent storing or revisiting past data~\cite{lesort2020continual}. Enabling robots to learn online, in a replay-free, single-pass manner, is therefore central to the long-standing goal of lifelong robot learning~\cite{silver2013lifelong,liu2017lifelong}.

This continuous streaming perspective differs fundamentally from standard lifelong imitation learning (LIL) formulations, which typically assume discrete task boundaries and repeated access to past trajectories~\cite{liu2023libero,wan2024lotus,lee2024incremental,yu2026lifelong}. Instead, a robot operating in the wild encounters a temporally correlated flow of multimodal data, including visual observations, language instructions, and proprioceptive feedback that rarely can be revisited ~\cite{thrun1995lifelong,jang2022bc}. Recent vision language action (VLA) models, spanning early systems such as RT-2~\cite{zitkovich2023rt} and Octo~\cite{team2024octo} to more recent large-scale policies (e.g., GR00T 1.5~\cite{bjorck2025gr00t}, $\pi_0.5$~\cite{intelligence2025vision}, $\pi_{0.7}$ ~\cite{intelligence2026pi}, SmolVLA~\cite{shukor2025smolvla}), illustrate the promise of pretrained backbones for general-purpose robotic control. Multi-task benchmarks like LIBERO~\cite{liu2023libero} further highlight their capacity for compositionality and transfer~\cite{liu2023tail,yu2026lifelong,hu2022lora}.
Yet these high-capacity representations tend to entangle instruction semantics and fine-grained context, making them fragile under sequential updates. In a single-pass setting where revisiting data is impossible, such entanglement leads to irreversible errors, a phenomenon also observed in continual learning with pretrained feature extractors~\cite{tiezzi2024memory,zhou2024continual}. More critically, single-pass learning demands rapid adaptation, yet such entanglement hinders the effective utilization of distinct knowledge, slowing router convergence and making it difficult to establish reliable routing decisions from a single exposure to incoming data.

The impact of representation entanglement is particularly pronounced in LIL, where the agent must integrate heterogeneous demonstrations while preserving previously acquired behaviors. This challenge becomes more severe when adaptation is performed on top of pretrained backbones, a common paradigm in modern visuomotor and vision-language-action (VLA) systems, where large models provide broad perceptual and linguistic priors before deployment. In such settings, sequential updates in a streaming regime can induce representation drift and catastrophic forgetting, progressively distorting the general-purpose features learned during pretraining~\cite{wan2024lotus,yu2026lifelong}.

Capturing both task semantics and context-dependent variations within a shared embedding further intensifies the plasticity–stability trade-off~\cite{kirkpatrick2017overcoming,kemker2018measuring,wang2024comprehensive}. In modular or expert-based architectures, this manifests as \textit{routing interference}, where learning new tasks shifts the representation space and disrupts previously established expert assignments, causing routing errors to accumulate over time in the absence of replay. Existing approaches based on parameter-efficient fine-tuning~\cite{hu2022lora,houlsby2019parameter,ding2023parameter,xin2024parameter} or dynamic expert libraries~\cite{lei2025dynamic} often rely on multi-epoch optimization or access to stored demonstrations to stabilize training, limiting their applicability in strict streaming regimes. In practice, these restrictive conditions are the norm. Human demonstrations are expensive and often cannot be stored due to privacy, memory, or deployment constraints~\cite{ravichandar2020recent,celemin2022interactive,zare2024survey}. A robot learning from a live stream of demonstrations must instead rely on online, replay-free, single-pass updates, where each trajectory provides a single opportunity for adaptation. Under these constraints, routing mechanisms must reliably disentangle instruction-level intent from rapidly shifting context, even as internal representations drift.

\begin{wrapfigure}{r}
{0.65\textwidth}
    \centering
\includegraphics[width=\linewidth, trim=1.cm .8cm 0.7cm .2cm, clip]{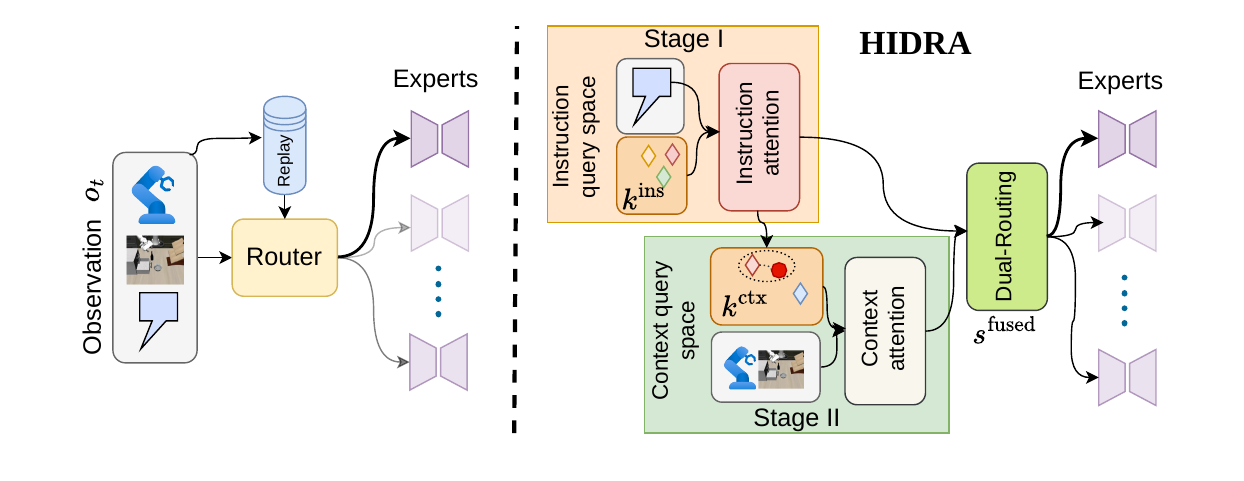}
    \caption{
    Left: Existing methods (e.g.,~\cite{lei2025dynamic}) perform expert routing using a single projection over entangled representations, which is unstable without replay.
    Right: \textsc{hidra} decouples routing into instruction retrieval and context refinement with regularization, ensuring reliable expert selection in replay-free LIL.}
    \label{fig:teaser} 
\end{wrapfigure}

To operate effectively in this setting, we introduce a Hierarchical Dual-Routing Attention mechanism (\textsc{hidra}) that decouples instruction-driven expert retrieval from context-driven refinement, reducing interference and enabling stable expert selection without replay. We additionally introduce a key-level geometric regularization strategy that structures the routing space and improves separability among experts under representational drift. Together, these components define a streaming LIL framework with an expandable expert set that learns from multimodal data streams without storing past demonstrations (see Fig.~\ref{fig:teaser}). We assume known task boundaries during lifelong adaptation, following the standard LIL protocol \cite{liu2023libero,wan2024lotus}; here, ``streaming'' refers to single-pass, non-revisitable task-wise data arrival rather than boundary-free task discovery.

\textbf{Our contributions are:} (i) a dual-routing attention mechanism that separates instruction semantics from contextual adaptation, enabling reliable expert selection under replay-free, single-pass learning; (ii) a key-level geometric regularization strategy that stabilizes routing by enforcing structure in the expert-selection space;
(iii) a unified streaming LIL framework with expandable expert libraries that achieves robust performance and forward transfer while maintaining low forgetting on diverse, long-horizon manipulation tasks with varied language instructions and object interactions, particularly under heterogeneous task sequences and language variation.

\section{Related Work}
\label{sec:relatedwork}

LIL studies how robotic policies can acquire new behaviors from sequentially arriving tasks while retaining previously learned skills under non-stationary task distributions \cite{hussein2017imitation,osa2018algorithmic,liu2023libero,wang2024comprehensive}. Existing approaches primarily differ in how they address the plasticity-stability trade-off \cite{kirkpatrick2017overcoming,kemker2018measuring}.

A first class of approaches relies on \emph{replay}, either by storing past demonstrations in a memory buffer or by generating pseudo-experience to approximate previous data distributions. Experience replay remains a strong baseline in continual learning \cite{chaudhry2019tiny}, and has been adopted in robotic lifelong learning through raw trajectory replay, retrieval-based adaptation, latent replay, and generative replay mechanisms \cite{gao2021cril,yang2024task,11217160,yu2026lifelong}. While effective, replay-based methods depend on access to past experience, and their cost grows with the size and complexity of stored data, making them less suitable under memory, privacy, or deployment constraints \cite{lesort2020continual,wang2024comprehensive}.

\emph{Regularization-based methods} preserve prior knowledge by penalizing harmful parameter updates or aligning new representations with previously learned ones \cite{kirkpatrick2017overcoming,roy2025m2distill}. While these methods avoid storing past data, they often struggle over long task sequences, where a single shared parameterization must simultaneously retain previous behaviors and adapt to new ones. In robotic LIL, multimodal distillation has been used to stabilize both latent representations and action outputs \cite{roy2025m2distill}; however, such methods operate in a largely monolithic model space and do not explicitly support selective reuse of modular knowledge.

\emph{Architectural approaches} instead aim to reduce interference by allocating or selecting task-relevant components over time. Task-conditioned methods isolate knowledge through pruning or task-specific adapters, often leveraging parameter-efficient fine-tuning (PEFT) techniques such as LoRA \cite{mallya2018packnet,hu2022lora,houlsby2019parameter,liu2023tail,zhu2025efficient, tiezzi2024continual, tiezzi2024memory}. While effective for retention, these methods typically assume access to task identity at inference time or restrict parameter sharing across tasks, limiting flexibility and forward transfer \cite{liu2023tail}. More recent observation-conditioned approaches maintain reusable skill or expert libraries and dynamically select components based on the current state or task context \cite{wan2024lotus,lee2024incremental,schmied2023learning,xu2025speci,zhang2025t2s,lei2025dynamic}. 
While these approaches enable modular reuse without task identifiers, they often rely on entangled representations to infer task identity and context simultaneously. This leads to routing ambiguity and interference, particularly in replay-free settings where maintaining expert separability is critical.

We build on modular and routing-based LIL, which facilitate knowledge reuse through the selection of task-specific components. LOTUS \cite{wan2024lotus} discovers reusable skills from demonstrations and composes them for continual manipulation, while ISCIL \cite{lee2024incremental} incrementally acquires retrievable skills through prototype-based organization and reusable adapters. 
Recent methods such as DMPEL \cite{lei2025dynamic} extend this paradigm by modulating pretrained models through learned selection over a shared representation space or an expanding set of reusable experts \cite{schmied2023learning,zhang2025t2s,xu2025speci}.
These methods are often evaluated in settings where replay, stored prototypes, or repeated optimization over past data stabilize routing decisions. However, in a replay-free, single-pass setting, such corrective signals are unavailable, making routing errors persistent and leading to increased interference as new tasks are introduced.
We adopt a strict single-pass, replay-free regime where routing cannot be corrected. Here, the primary challenge is \emph{routing interference}, where new components disrupt the selection of previously learned experts. \textsc{hidra} mitigates this by decoupling task identity from context through hierarchical, regularized routing, ensuring stable modular reuse without revisiting past data.

\section{Method}
\label{sec:method}
We introduce Hierarchical Dual-Routing Attention (\textsc{HIDRA}) for replay-free, single-pass lifelong imitation learning (LIL). Our method tackles the core limitation of expert-based LIL: routing interference. Without replay, representation drift leads standard routers to activate experts from unrelated tasks. \textsc{HIDRA} prevents this by splitting routing into a stable, instruction-guided retrieval stage and a context-dependent refinement stage. The agent is first trained via multi-task behavioral cloning (BC) and then adapted to a stream of unseen tasks, each observed only once and never stored.

\begin{figure}
    \centering
    \includegraphics[width=\linewidth, trim=1.6cm 1.7cm 14cm 3.5cm, clip]{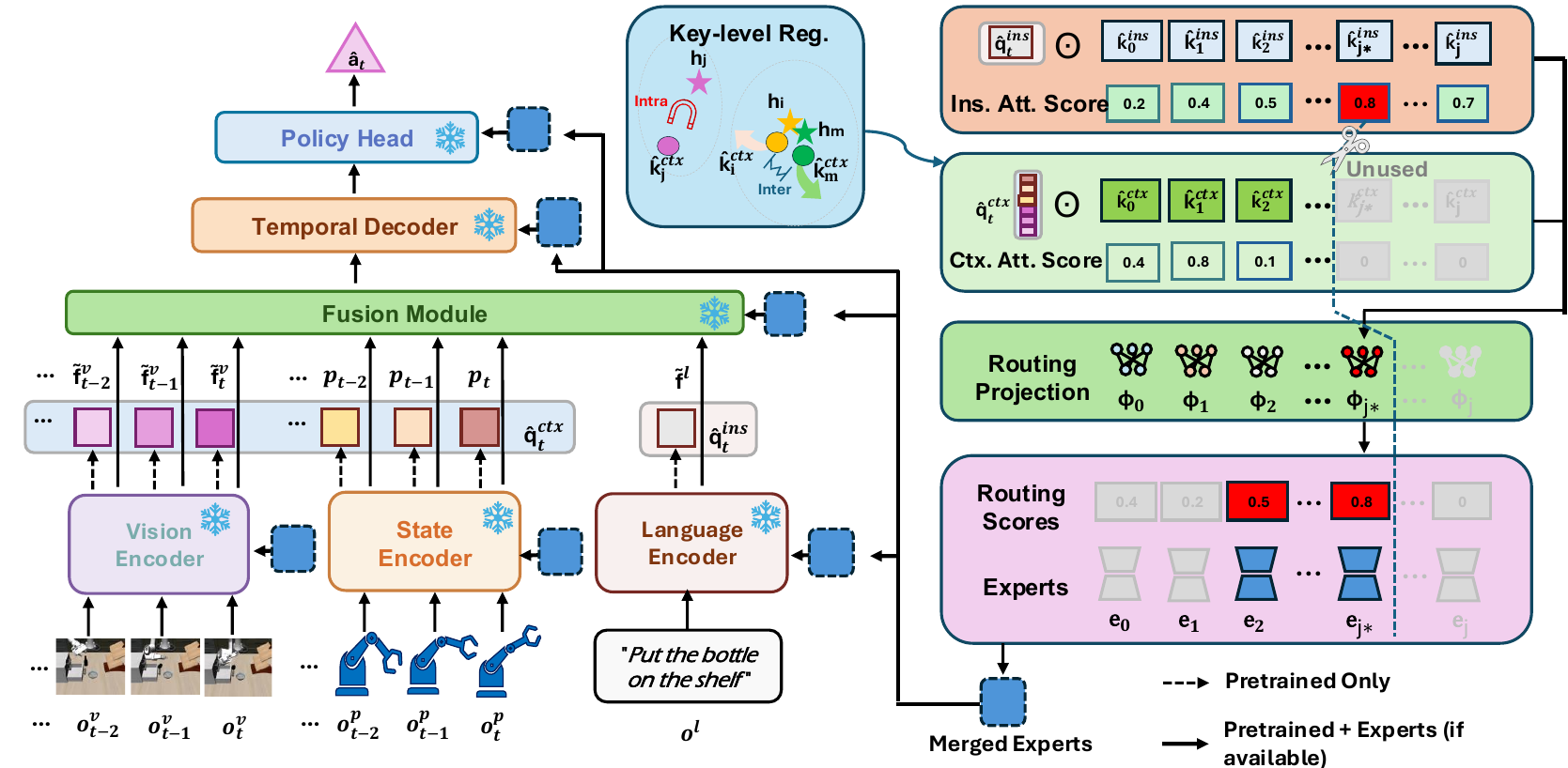} 
    \caption{Overview of \textsc{hidra}. A frozen pretrained multimodal policy produces an instruction query $q_t^{\mathrm{ins}}$ and a context query $q_t^{\mathrm{ctx}}$. These drive a two-stage hierarchical dual-routing attention mechanism. 
Stage~I performs instruction-level retrieval via $q_t^{\mathrm{ins}}$ to select an anchor expert $j^*$ and an instruction candidate set $\mathcal{S}_{j^*}$ over task-level keys $\{k_j^{\mathrm{ins}}\}$. 
Stage~II then uses $q_t^{\mathrm{ctx}}$ to perform context-level refinement within $\mathcal{S}_{j^*}$ using context keys $\{k_j^{\mathrm{ctx}}\}$, producing fused routing scores and selecting the active experts. 
Key-space regularization (Reg.) promotes intra-task compactness and inter-task separation, enabling stable and interference-resistant expert selection as new task-specific experts are introduced during replay-free lifelong adaptation.}
    \label{fig:main_method}
\end{figure}

\subsection{Preliminaries}

\textbf{Pre-Training and Lifelong Adaptation.} We consider replay-free lifelong imitation learning as a two-stage process. 
During pre-training, the agent is trained on a set of tasks 
$\mathcal{T}^{\mathrm{pre}}=\{T'_j\}_{j=1}^{J}$, where each task $T'_j$ provides demonstrations 
$\mathcal{D}'_j=\{\tau_{j,i}\}_{i=1}^{N_j}$ and each trajectory 
$\tau_{j,i}=\{(o_t,a_t)\}_{t=1}^{L_j}$ contains multimodal observations $o_t$ 
(images, proprioception, and language instruction) and expert actions $a_t$. 
We learn a policy $\pi_\theta$ by minimizing the multi-task behavioral cloning objective 
$\min_{\theta}\sum_{j=1}^{J}\mathbb{E}_{(o_{\le t},a_t)\sim\mathcal{D}'_j}
\mathcal{L}_{\mathrm{BC}}\!\left(\pi_\theta(\cdot \mid o_{\le t}),a_t\right)$,
where $o_{\le t}$ denotes the recent observation history and $\mathcal{L}_{\mathrm{BC}}$ is a supervised imitation loss, e.g., negative log-likelihood. 
After pre-training, the agent observes a sequence of novel tasks 
$\mathcal{T}^{\mathrm{LL}}=\{T_1,\dots,T_K\}$, disjoint from $\mathcal{T}^{\mathrm{pre}}$ but sharing the same observation and action spaces. 
At lifelong task $k$, demonstrations $\mathcal{D}_k$ are available only once and are not stored; therefore, we adapt only a subset of parameters $\theta^\star$ using 
$\min_{\theta^\star}\mathbb{E}_{(o_{\le t},a_t)\sim\mathcal{D}_k}
\mathcal{L}_{\mathrm{BC}}\!\left(\pi_{\theta^\star}(\cdot \mid o_{\le t}),a_t\right)$.


\textbf{Frozen multimodal backbone.}
The policy follows a standard vision-language-action architecture with modality-specific encoders, a fusion module, a temporal decoder, and an action head. At each time step, the multimodal observation is denoted as
$o_t=(o_t^v,o_t^p,o^l)$, where $o_t^v$ is the visual observation, $o_t^p$ is the proprioceptive state, and $o^l$ is the language instruction shared across the trajectory, i.e., $o_t^l=o^l$ for all $t$.
Pre-trained backbone parameters remain frozen during adaptation.

\textbf{Expandable LoRA expert library.}
We attach lightweight low-rank experts to a set of selected linear layers of the frozen backbone.
Consider one such linear layer with input $x\in\mathbb{R}^{d_{\mathrm{in}}}$ and frozen parameters
$W_0\in\mathbb{R}^{d_{\mathrm{out}}\times d_{\mathrm{in}}}$ and $b_0\in\mathbb{R}^{d_{\mathrm{out}}}$.
For each new task $r$, we instantiate a new set of LoRA adapters $\{A_r, B_r\}$ at every selected layer. During the training of task $r$, only these newly created adapters and their corresponding routing parameters are optimized, while those from previous tasks $\{1, \dots, r-1\}$ remain frozen. 
At each timestep $t$, the output of the augmented layer is $
y_t = (W_0 x_t + b_0) + \sum_{j=1}^{r} 2 *c_{j,t} (B_j A_j x_t + b_j)
$,
where $c_{j,t} \in [0, 1]$ are dynamic routing coefficients. This structure ensures that each task is represented by a distributed "expert" composed of adapters across the entire architecture. This parameter isolation allows capacity to grow incrementally while preventing the degradation of previously acquired skills, as past experts receive no gradient updates. Notably, routing is computed independently at each LoRA-adapted layer; for clarity, we describe one layer and omit its index.


\subsection{\textsc{hidra}: Hierarchical Dual-Routing Attention}
\label{subsec:dual_routing}

Existing routing mechanisms often rely on a single entangled representation, which must simultaneously encode task identity and momentary context \cite{lei2025dynamic}.
This coupling makes expert selection sensitive to representation drift, especially in replay-free lifelong imitation learning, where past data cannot be revisited to stabilize the router.
To address this issue, \textsc{hidra} decouples routing into two stages performed via dot-product attention: (i) instruction-level retrieval, which uses stable task semantics to identify an anchor expert and thereby defines a restricted pool of previously learned experts for subsequent context refinement, and (ii) context-level refinement, which uses the current observation to select relevant experts only within that restricted set.

Specifically, in \textsc{hidra} each expert $e_j$ maintains two learnable keys: an instruction key
$k_j^{\mathrm{ins}} \in \mathbb{R}^{d_k}$, initialized from the corresponding instruction representation, and a context key
$k_j^{\mathrm{ctx}} \in \mathbb{R}^{d_s}$, learned to capture state-dependent relevance.
We define the instruction query $q_t^{\mathrm{ins}}$ and the context query $q_t^{\mathrm{ctx}}$ as:
$q_t^{\mathrm{ins}} = g_{\mathrm{pool}}(\tilde{f}^{\ell}_{t-T:t}) \in \mathbb{R}^{d_k}, \quad q_t^{\mathrm{ctx}} = g_{\mathrm{pool}}([\tilde{f}^{v}_{t-T:t}, p_{t-T:t}]) \in \mathbb{R}^{d_s}$,
where $\tilde{f}^{\ell}$, $\tilde{f}^{v}$, and $p$ represent the frozen language, visual, and proprioceptive embeddings, respectively. Here, the transformation $g_{\mathrm{pool}}(\cdot)$ denotes a temporal mean-pooling operation over the $T$-step window, which serves as a stable, non-parametric projection into the key space. This decoupling ensures that $q_t^{\mathrm{ins}}$ captures the invariant task semantics while $q_t^{\mathrm{ctx}}$ focuses exclusively on the momentary visual and physical state. In our implementation, all routing parameters associated with a task are introduced and optimized only during that task's learning phase and are subsequently frozen to prevent interference, adhering to the expert-isolation principle. We apply $\ell_2$-normalization to the queries and keys before attention score computation, and use $\hat{q}$ and $\hat{k}$ to denote the normalized vectors.

\textbf{Stage I: Instruction-level retrieval.}
The first stage performs coarse expert retrieval using only the task instruction.
This stage provides a stable semantic anchor before any context-dependent refinement is performed, which is crucial in replay-free LIL because context representations may drift as new tasks are learned.
For each expert $e_j$, we compute an instruction score
$\
a^{\mathrm{ins}}_{j,t}
=
\max\!\left(
0,\,
\langle \hat{q}_t^{\mathrm{ins}}, \hat{k}_j^{\mathrm{ins}} \rangle
/ \tau_{\mathrm{ins}}
\right)$,
\
where $\langle \cdot, \cdot \rangle$ denotes the dot product, and
$\tau_{\mathrm{ins}}>0$ is a temperature controlling score sharpness.
Then, an anchor expert is selected as
$j^* = \arg\max_{j \in \{1,\dots,r\}} a^{\mathrm{ins}}_{j,t}$.
The anchor expert provides a task-level reference determined by language semantics.
To reduce unrestricted competition over the expanding expert library, we use the instruction-selected anchor to define a restricted candidate pool for context refinement. Concretely, we form an anchor-conditioned candidate set $\mathcal{S}_{j^*}$ consisting of experts introduced before the selected anchor, $\mathcal{S}_{j^*} = \{j \in \{1, \dots, r\} \mid j < j^*\}$. This set represents the pool of earlier acquired knowledge available for potential reuse in Stage II. Thus, the instruction branch provides the semantic anchor and instruction-level compatibility scores, whereas the context branch determines which experts to reuse within this restricted pool.

\textbf{Stage II: Context-level refinement.}
Within the anchor-conditioned candidate set $\mathcal{S}_{j^*}$, we compute a context score for each $j \in \mathcal{S}_{j^*}$ as
$a^{\mathrm{ctx}}_{j,t} = \langle \hat{q}_t^{\mathrm{ctx}}, \hat{k}_j^{\mathrm{ctx}} \rangle / \tau_{\mathrm{ctx}}$,
with temperature $\tau_{\mathrm{ctx}}>0$.
For each candidate expert, we combine instruction-level and context-level evidence into a scalar fused routing \emph{score}
$s^{\mathrm{fused}}_{j,t} = \phi_j(a^{\mathrm{ins}}_{j,t}, a^{\mathrm{ctx}}_{j,t})$,
which integrates task-level compatibility and state-dependent relevance after the instruction branch has restricted the set of eligible experts. We define $K^*$ = min($K-1$, $j^*-1$).
The final context-refined experts are selected by applying $Top-K^*$ over fused routing scores inside the anchor-conditioned candidate set:
$\mathcal{T}_t
=
\arg\, Top-K^*_{j \in \mathcal{S}_{j^*}}
s^{\mathrm{fused}}_{j,t}$,
where $Top-K^*$ returns the indices of the $K^*$ largest values.
Therefore, the selected experts are not simply those with the highest context scores over the whole library; they are experts that are both instruction-consistent and contextually relevant.
This is important: the instruction branch determines which experts are allowed to compete, while the context branch refines the selection only within this safe subset. Notably, experts with $j > j^*$ are excluded because Stage II is defined to reuse experts learned before the anchor. Allowing experts introduced later to modify selection for an earlier anchor would change its routing without access to the earlier task data, which are unavailable in the replay-free setting. 

\textbf{Routing coefficients and their role in the LoRA mixture.}
The final routing coefficients combine the instruction anchor with the selected context-refined experts. 
The routing coefficients are  defined as
$c_{j,t} = a^{\mathrm{ins}}_{j,t}\,{1}[j=j^*]
+ s^{\mathrm{fused}}_{j,t}\,{1}[j \in \mathcal{T}_t]$, where ${1}[\cdot]$ denotes the indicator function.  

The anchor expert is weighted directly by its instruction score, ensuring that the current routing decision remains grounded in stable semantic information.
The additional selected experts are weighted by their fused routing scores, allowing \textsc{hidra} to reuse previously learned skills only when both instruction-level and context-level evidence support their activation. 

Overall, \textsc{hidra} enforces sparse and structured expert selection.
Stage I identifies a stable instruction-driven anchor and restricts the search space to an anchor-conditioned candidate set.
Stage II then performs context-dependent refinement only within this set using fused instruction--context scores.
As a result, \textsc{hidra} avoids routing interference by preventing unrelated experts from entering the context-dependent competition, while still retaining the flexibility to reuse useful prior skills under the restricted pool.

\subsection{Key-Space Regularization}
\label{subsec:key_reg}
Because routing relies on similarity in the key space, we regularize keys to maintain a stable geometry as the library grows.
We regularize context keys $\{k_j^{\mathrm{ctx}}\}$, which govern the fine-grained refinement.

\textbf{Per-task reference prototypes.}
For each expert $e_j$ created for task $k$, we maintain a reference prototype $h_j \in \mathbb{R}^{d_s}$ that summarizes the context state distribution of that task. 
During the training phase of task $k$, $h_j$ is computed as the cumulative running average of all normalized context queries observed so far. 
Specifically, at each update step, we maintain a counter $n$ of the number of observed queries and update $h_j$ as:
\begin{equation}
h_j \leftarrow \frac{n-1}{n} h_j + \frac{1}{n} \hat{q}_t^{\mathrm{ctx}}, 
\qquad 
h_j \leftarrow \frac{h_j}{\|h_j\|_2}
\label{eq:prototype}
\end{equation}
where $n$ denotes the number of context queries accumulated for the current task. 

\textbf{Intra-task alignment.}
We encourage each context key to align with its prototype:
$\mathcal{L}_{\mathrm{intra}} = \sum_{j \in \mathcal{J}_k} \Big(1 - \langle \hat{k}_j^{\mathrm{ctx}}, h_j \rangle \Big)$,
where $\mathcal{J}_k$ indexes experts belonging to the current task, 
and $\langle \cdot, \cdot \rangle$ denotes the inner product. This loss maximizes similarity between each context key and its task prototype, encouraging keys to encode task-relevant semantics and improving routing.

\textbf{Inter-task separation.}
To improve discriminability between tasks, we penalize similarity between context keys from different tasks: $\mathcal{L}_{\mathrm{inter}} = \sum_{j \in \mathcal{J}_j} \sum_{i \in \mathcal{I}_{<j}} \langle \hat{k}_j^{\mathrm{ctx}}, \hat{k}_i^{\mathrm{ctx}} \rangle^2$,
where $\mathcal{I}_{<j}$ indexes experts created before task $k$. 
This encourages approximate orthogonality across tasks while allowing within-task structure.

\textbf{Lifelong objective.}
During task $k$, we optimize the BC loss on $\mathcal{D}_k$ together with key regularization:
$\mathcal{L} = \mathcal{L}_{\mathrm{BC}} + \lambda_{\mathrm{intra}}\,\mathcal{L}_{\mathrm{intra}} + \lambda_{\mathrm{inter}}\,\mathcal{L}_{\mathrm{inter}}$
with hyperparameters $\lambda_{\mathrm{intra}},\lambda_{\mathrm{inter}}\ge 0$.

\textbf{Summary.}
Our method replaces router-based coefficient prediction (and its replay stabilization) with sparse attention over \emph{learnable} expert keys.
Decoupling instruction retrieval from context refinement yields a structured, drift-robust routing mechanism, while key-space regularization preserves separability as the expert library expands. Additional implementation details (e.g., decoder type, expert configuration, and projection layers) are provided in Appendix~\ref{sec:imple}.

\section{Experimental Analysis}
\label{sec:experiments}

\textbf{Benchmark suites.} We evaluate our method on the LIBERO lifelong robotic manipulation benchmark~\cite{liu2023libero}, which includes diverse tasks (e.g., turning on a stove, moving cups, opening drawers). Following prior works~\cite{lei2025dynamic,liu2023libero,yu2026lifelong,liu2023tail}, we use four standard suites: LIBERO-OBJECT, LIBERO-GOAL, LIBERO-SPATIAL, and LIBERO-10 (also called LIBERO-LONG), covering object-centric reasoning, goal-directed behavior, spatial understanding, and long-horizon manipulation.
To study more challenging conditions, we introduce LIBERO-MIX, constructed by sampling five tasks from each suite and interleaving them into a longer, more heterogeneous sequence. This increases task diversity and induces frequent distribution shifts, amplifying cross-task interference in replay-free, single-pass learning. Tasks are sampled uniformly from each suite (full list in the Appendix), resulting in kitchen-centric manipulation scenarios with consistent environments but diverse task semantics.
All datasets provide multimodal observations, including visual inputs (\textit{agent-view} and \textit{eye-in-hand}), language instructions, robot state (e.g., joint positions, velocities, gripper status), and actions, enabling visually grounded, language-conditioned policy learning. 
During pre-training, we use LIBERO-90 with 50 demonstrations per task as in~\cite{liu2023libero,lei2025dynamic}. In the lifelong phase, tasks are introduced sequentially (10 tasks per benchmark, 50 demonstrations each). For LIBERO-MIX, we consider \textit{Inc4} (four tasks per step) and \textit{Inc1} (fully sequential); Inc4 is more challenging due to increased interference. Unless specified otherwise, all methods are trained for a single epoch (single-pass). Multi-pass results are reported in the Appendix~\ref{secAppen:MultiPass}.

\textbf{Methods under comparison.}  
We compare against several state-of-the-art (SOTA) methods, including replay-based and replay-free approaches. Replay-based methods include Experience Replay (ER)~\cite{liu2023libero,chaudhry2019tiny}, which stores raw trajectories; MLR-IFA~\cite{yu2026lifelong}, which replays latent multimodal features; and DMPEL~\cite{lei2025dynamic}, which relies on replay for retention and uses a non-attentional routing mechanism. For DMPEL, we evaluate different replay ratios: DMPEL (100\%), which replays all past data, and DMPEL (50\%), which replays a randomly sampled half at each update, enabling analysis of replay quantity.
Replay-free methods include L2M~\cite{schmied2023learning}, EWC~\cite{kirkpatrick2017overcoming}, ISCIL~\cite{lee2024incremental}, and TAIL~\cite{liu2023tail}. We also report a replay-free variant, DMPEL (0\%), for direct comparison. All methods, including our \textsc{HIDRA}, share the same pretrained initialization~\cite{lei2025dynamic} to ensure a fair comparison.

\textbf{Evaluation metrics.} We use standard lifelong learning metrics~\cite{wan2024lotus,rodriguez2018don,liu2023libero}: Forward Transfer (FWT), Negative Backward Transfer (NBT), and Area Under the Curve (AUC), computed from task success rates~\cite{liu2023libero}. We follow prior work~\cite{wan2024lotus,liu2023libero,lee2024incremental,yu2026lifelong} for definitions.
Let $M$ be the number of tasks and $r_{i,j}$ the success rate on task $j$ after learning $i$ tasks. FWT measures adaptation: $\mathrm{FWT} = \frac{1}{M} \sum_{m=1}^{M} r_{m,m}$. NBT measures forgetting: $\mathrm{NBT}_m = \frac{1}{M-m} \sum_{q=m+1}^{M} (r_{m,m} - r_{q,m})$, and $\mathrm{NBT} = \frac{1}{M-1} \sum_{m=1}^{M-1} \mathrm{NBT}_m$. AUC measures overall performance: $\mathrm{AUC} = \frac{1}{M} \sum_{m=1}^{M} \mathrm{AUC}_m$, where $\mathrm{AUC}_m = \frac{1}{M-m+1} \left(r_{m,m} + \sum_{q=m+1}^{M} r_{q,m}\right)$.

\textbf{Setup.}
Results are averaged over three random seeds. Each policy is evaluated on 20 trials per seed with different environment initializations (e.g., gripper states). All experiments are conducted on a single NVIDIA A100 GPU.
Following~\cite{lei2025dynamic,liu2023tail}, pre-training uses a batch size of 8 for 2 epochs. During lifelong learning, we use a batch size of 32 and optimize with AdamW (learning rate $1\times10^{-4}$, $(\beta_1, \beta_2) = (0.9, 0.999)$). Regularization weights are set to $\lambda_{\mathrm{intra}}=1$ and $\lambda_{\mathrm{inter}}=1$. The instruction and context temperatures are set to $\tau_{\mathrm{ins}} = d_k$ and $\tau_{\mathrm{ctx}} = d_s$, respectively, following~\cite{lei2025dynamic}. In the final routing stage, we select $K=3$ experts.

\subsection{Results}
\label{sec:results}

We compare against SOTA and present key ablations; additional analyses are deferred to the Appendix, including extended ablations, routing and representation studies, efficiency evaluations, robustness analyses (e.g., multi-pass training), full benchmark details, statistical significance analysis, and a discussion of limitations.

\begin{table*}[tb!]
\centering
\caption{Comparison on LIBERO benchmarks (mean $\pm$ std). Best shown in \textbf{bold.}}
\label{tab:libero_main}
\resizebox{\textwidth}{!}{%
\begin{tabular}{lcccccccccccccc}
\toprule
\textbf{Method} & \textbf{Replay} &
\multicolumn{3}{c}{\textbf{LIBERO-OBJECT}} &
\multicolumn{3}{c}{\textbf{LIBERO-GOAL}} &
\multicolumn{3}{c}{\textbf{LIBERO-SPATIAL}} &
\multicolumn{3}{c}{\textbf{LIBERO-10}} \\
\cmidrule(lr){3-5}\cmidrule(lr){6-8}\cmidrule(lr){9-11}\cmidrule(lr){12-14}
 & \textbf{Free} & FWT$\uparrow$ & NBT$\downarrow$ & AUC$\uparrow$
 & FWT$\uparrow$ & NBT$\downarrow$ & AUC$\uparrow$
 & FWT$\uparrow$ & NBT$\downarrow$ & AUC$\uparrow$
 & FWT$\uparrow$ & NBT$\downarrow$ & AUC$\uparrow$ \\
\midrule

ER \cite{liu2023libero,chaudhry2019tiny} & \xmark
& 46.0$\pm$3.6 & 53.3$\pm$3.2 & 19.8$\pm$2.9
& 36.0$\pm$2.6 & 17.3$\pm$3.4 & 23.6$\pm$3.3
& 15.0$\pm$2.8 & -1.8$\pm$4.5 & 16.2$\pm$3.7
& 7.0$\pm$2.9 & 4.7$\pm$4.0 & 3.4$\pm$3.5 \\

MLR-IFA \cite{yu2026lifelong} & \xmark
& 34.0$\pm$2.0 & \textbf{-6.2$\pm$2.4} & 38.7$\pm$2.2
& 43.0$\pm$2.4 & 17.3$\pm$1.6 & 38.2$\pm$2.1
& 27.0$\pm$1.5 & 10.1$\pm$1.9 & 19.9$\pm$2.0
& 21.1$\pm$2.7 & \textbf{-4.9$\pm$3.0} & 24.6$\pm$2.5 \\

DMPEL (50\%) \cite{lei2025dynamic} & \xmark
& 77.0$\pm$1.2 & -4.4$\pm$0.6 & 78.0$\pm$1.6
& 62.0$\pm$1.8 & 4.2$\pm$1.5 & 61.0$\pm$1.7
& 46.0$\pm$2.9 & 2.7$\pm$2.6 & 43.8$\pm$2.0
& \textbf{62.0$\pm$1.5} & 7.9$\pm$3.0 & 54.5$\pm$1.5 \\

DMPEL (100\%) \cite{lei2025dynamic} & \xmark
& 82.0$\pm$1.0 & -0.7$\pm$0.8 & 82.2$\pm$1.4
& 64.0$\pm$1.6 & \textbf{-0.4$\pm$1.4} & 64.3$\pm$1.8
& 44.0$\pm$2.6 & \textbf{-4.3$\pm$1.2} & 47.0$\pm$2.3
& 56.5$\pm$2.6 & 4.7$\pm$0.3 & 53.8$\pm$2.6 \\

\midrule

L2M \cite{schmied2023learning} & \cmark
& 1.0$\pm$0.5 & 0.9$\pm$0 & 0.7$\pm$0.2
& 4.0$\pm$0.6 & 2.8$\pm$0.5 & 2.3$\pm$0
& 8.0$\pm$0.6 & 3.1$\pm$0.4 & 5.6$\pm$0.4
& 7.0$\pm$0.8 & 4.5$\pm$0.3 & 3.7$\pm$1.1 \\

EWC \cite{kirkpatrick2017overcoming} & \cmark
& 52.0$\pm$2.5 & 35.4$\pm$3.0 & 28.3$\pm$1.8
& 27.0$\pm$2.2 & 16.3$\pm$1.6 & 14.6$\pm$1.4
& 20.0$\pm$1.8 & 16.6$\pm$2.6 & 7.3$\pm$1.7
& 17.0$\pm$1.5 & 6.2$\pm$0.8 & 12.0$\pm$3.1 \\

TAIL \cite{liu2023tail} & \cmark
& 50.0$\pm$1.2 & 0.0$\pm$0.0 & 50.0$\pm$1.2
& 25.0$\pm$1.4 & 0.0$\pm$0.0 & 25.0$\pm$1.4
& 18.0$\pm$1.5 & 0.0$\pm$0.0 & 18.0$\pm$1.5
& 16.0$\pm$1.8 & 0.0$\pm$0.0 & 16.0$\pm$1.8 \\

ISCIL \cite{lee2024incremental} & \cmark
& 39.0$\pm$1.3 & -3.7$\pm$2.8 & 41.2$\pm$1.6
& 31.0$\pm$1.2 & 1.6$\pm$0.4 & 28.9$\pm$0.8
& 14.5$\pm$0.7 & -0.9$\pm$0.3 & 14.6$\pm$1.3
& 19.5$\pm$1.3 & 9.5$\pm$0.6 & 18.8$\pm$1.2 \\

DMPEL (0\%) \cite{lei2025dynamic} & \cmark
& 73.0$\pm$2.6 & 73.3$\pm$5.8 & 24.7$\pm$4.4
& 61.0$\pm$3.1 & 72.2$\pm$5.6 & 23.1$\pm$4.1
& 49.0$\pm$2.5 & 49.4$\pm$3.6 & 13.9$\pm$2.9
& 47.0$\pm$3.3 & 46.4$\pm$4.0 & 11.7$\pm$3.5 \\

\rowcolor{babyblue}
\textsc{hidra} (Ours) & \cmark
& \textbf{93.0$\pm$1.5} & 2.2$\pm$1.9 & \textbf{91.5$\pm$0.1}
& \textbf{67.5$\pm$2.2} & 0.9$\pm$5.5 & \textbf{66.7$\pm$2.0}
& \textbf{58.0 $\pm$1.5} & -1.4 $\pm$0.8 & \textbf{58.7 $\pm$1.0}  
& 58.2$\pm$1.5 & 0.1$\pm$3.1 & \textbf{58.1$\pm$1.5} \\
\bottomrule
\end{tabular}}
\end{table*}

\subsubsection{Comparison with SOTA Methods}
As shown in Tab.~\ref{tab:libero_main}, \textsc{hidra} achieves the best AUC across all four benchmarks, outperforming prior methods. It achieves near-zero NBT in most settings, with negative values in some cases, indicating that previously learned tasks are largely retained without replay. Compared to strong SOTA such as DMPEL (100\%), \textsc{hidra} consistently attains higher AUC while maintaining low forgetting. These improvements are particularly evident on more challenging benchmarks such as LIBERO-SPATIAL and LIBERO-10, where \textsc{hidra} also exhibits strong forward transfer. These results further suggest that our proposed routing mechanism enables faster convergence to reliable expert assignments, facilitating effective adaptation under the single-pass constraint.

\begin{table*}[tb!]
\centering
\caption{Comparison of results on the LIBERO-MIX and LIBERO-INS benchmarks (mean $\pm$ std). Best shown in \textbf{bold.} }
\label{tab:libero_mix}
\resizebox{\textwidth}{!}{%
\begin{tabular}{lccccccccc}
\toprule
\multirow{2}{*}{} &
\multicolumn{3}{c}{LIBERO-MIX @ Inc1} &
\multicolumn{3}{c}{LIBERO-MIX @ Inc4} &
\multicolumn{3}{c}{LIBERO-INS} \\
\cmidrule(lr){2-4}\cmidrule(lr){5-7}\cmidrule(lr){8-10}
& FWT$\uparrow$ & NBT$\downarrow$ & AUC$\uparrow$
& FWT$\uparrow$ & NBT$\downarrow$ & AUC$\uparrow$
& FWT$\uparrow$ & NBT$\downarrow$ & AUC$\uparrow$ \\
\midrule

MLR-IFA \cite{yu2026lifelong}
& 28.5 $\pm$3.4 & 4.5 $\pm$ 2.0 &  25.0$\pm$ 2.1
& 21.5 $\pm$ 3.0 & \textbf{-5.2 $\pm$ 2.2} & 24.7 $\pm$ 2.8 
& 4.0 $\pm$ 1.8 & -4.9 $\pm$ 6.2 & 7.9 $\pm$ 2.6 \\

DMPEL (100\%) \cite{lei2025dynamic}
& 53.0 $\pm$ 1.0 & \textbf{-3.9 $\pm$ 2.8} & 55.2 $\pm$ 2.5
& 61.0 $\pm$ 1.7 & 2.4 $\pm$ 0.9 & 59.9 $\pm$ 1.4 
& 23.0 $\pm$ 3.6 & \textbf{-6.1 $\pm$ 8.0} & 27.1 $\pm$ 4.3 \\

ISCIL \cite{lee2024incremental}
& 21.1 $\pm$ 2.9 & 0.4 $\pm$ 1.6 & 20.6 $\pm$ 2.8
& 14.8 $\pm$ 3.5 & -0.3 $\pm$ 1.3 & 14.7 $\pm$ 0.4
& 11.0 $\pm$ 2.7 & 1.7 $\pm$ 2.1 & 10.0 $\pm$ 3.2 \\

\rowcolor{babyblue}
\textsc{hidra} (Ours)
& \textbf{71.3 $\pm$ 2.1} & 2.8 $\pm$ 1.7 & \textbf{68.0 $\pm$ 1.0}
& \textbf{66.2 $\pm$ 2.5} & -2.0 $\pm$ 3.2 & \textbf{67.0 $\pm$ 1.2}
& \textbf{25.5 $\pm$ 1.9} & -3.0 $\pm$ 2.4 & \textbf{27.6 $\pm$ 3.8} \\

\bottomrule
\end{tabular}
}
\end{table*}

Tab.~\ref{tab:libero_mix} reports results in more challenging settings against the strongest SOTA from Tab.~\ref{tab:libero_main}. On LIBERO-MIX, \textsc{hidra} achieves the best FWT and AUC in both Inc1 and Inc4, with larger gains in Inc4 where multiple tasks are introduced simultaneously, indicating stronger robustness to task interference. Forgetting remains competitive, outperforming DMPEL and ISCIL in Inc4. On LIBERO-INS, \textsc{hidra} achieves the best overall performance, combining higher AUC with competitive forgetting. These results indicate that the proposed routing mechanism yields a better balance between adaptation and retention in more demanding scenarios.

\subsubsection{Ablation Studies}
\label{sec:ablation}
We conduct a series of ablation experiments to analyze the key components of \textsc{HIDRA}. In particular, we study the effect of the routing design, the contribution of intra- and inter-task regularization, different prototype estimation strategies, alternative inter-task regularization objectives, and routing combination strategies. 

\textbf{Ablation of routing and regularization components.}
We consider a simple routing baseline in which expert selection is performed using a single projection module without attention or explicit separation between task-level and context-level signals. As shown in Tab.~\ref{tab:ablation_all}, this baseline performs poorly, with high NBT and low AUC, highlighting the limitations of unstructured routing in replay-free settings.
Using a single attention branch improves performance, with instruction-based routing outperforming context-only routing, indicating that task-level signals provide stronger guidance for expert selection. However, single-branch routing remains limited, as it cannot jointly capture task identity and state-dependent variations. 
We then introduce dual-attention routing without regularizing the context keys. In this case, after selecting the instruction-guided adapter, the reuse of previously learned adapters in $\mathcal{S}_{j^*}$ is unstructured and close to random. This further improves performance, demonstrating the benefit of reusing previously learned adapters even when the context-based reuse is effectively close to random.
Finally, we incorporate intra-task and inter-task regularization to further control the reuse of past adapters. This further improves AUC and, in most cases, FWT, with the combination of both achieving the best overall performance on these metrics. Improvements in NBT are less consistent, but the results suggest complementary effects on routing behavior. Together, these components lead to more structured routing behavior and improved overall performance.

\begin{table}[tb!]
\centering
\caption{
Ablation of the routing and regularization components of the proposed method.}
\label{tab:ablation_all}
\resizebox{0.9\linewidth}{!}{
\begin{tabular}{lcccccc}
\toprule
\textbf{} & \multicolumn{3}{c}{\textbf{LIBERO-OBJECT}} & \multicolumn{3}{c}{\textbf{LIBERO-SPATIAL}} \\
\cmidrule(lr){2-4} \cmidrule(lr){5-7}
 & FWT$\uparrow$ & NBT$\downarrow$ & AUC$\uparrow$ & FWT$\uparrow$ & NBT$\downarrow$ & AUC$\uparrow$ \\
\midrule

Simple Routing (no attention) 
& 71.0 $\pm$ 3.7 & 62.0 $\pm$ 2.9 & 26.1 $\pm$ 2.5
& 43.5 $\pm$2.5 & 47.1 $\pm$3.8 & 9.2 $\pm$2.0  \\

$\text{SingleAttn}(a^{\text{ctx}}_{j,t})$  
& 72.0 $\pm$ 1.9 & 2.5 $\pm$ 2.1 & 70.0 $\pm$ 2.2
& 41.0 $\pm$ 3.4 & -0.3 $\pm$ 2.6 & 41.1 $\pm$ 3.1  \\

$\text{SingleAttn}(a^{\text{ins}}_{j,t})$  
& 83.0 $\pm$ 2.4 & -0.5 $\pm$ 1.2 & 83.2 $\pm$ 2.0
& 53.0 $\pm$1.6 & 3.0 $\pm$0.7 & 51.7 $\pm$1.6 \\

$\text{DualAttn}(a^{\text{ins}}_{j,t}, a^{\text{ctx}}_{j,t})$ 
 & 85.5 $\pm$ 3.1 & -1.0$\pm$ 2.2 & 85.8$\pm$ 2.6 
 & 54.5 $\pm$1.7 & 1.7 $\pm$0.7 & 54.2 $\pm$1.9  \\

$+\ \mathcal{L}_{\text{intra}}$ 
& 87.5 $\pm$ 2.2 & \textbf{-1.9 $\pm$ 1.3} & 87.4 $\pm$ 2.6
& 55.5 $\pm$2.9 & 2.1$\pm$2.7 & 54.6$\pm$3.3 \\

$+\ \mathcal{L}_{\text{inter}}$ 
& 90.0 $\pm$1.8 & 1.2 $\pm$1.3 & 88.8 $\pm$0.9
& 57.0$\pm$2.0 & 1.2$\pm$4.4 & 55.8$\pm$2.6 \\

\rowcolor{babyblue}
$+\ \mathcal{L}_{\text{intra}} + \mathcal{L}_{\text{inter}}$ (Ours)
& \textbf{93.0 $\pm$1.5} & 2.2 $\pm$1.9 & \textbf{91.5 $\pm$0.1}
& \textbf{58.0 $\pm$1.5} & \textbf{-1.4 $\pm$0.8} & \textbf{58.7 $\pm$1.0}  \\

\bottomrule
\end{tabular}}
\end{table}

\begin{figure*}[t]
    \centering
    \begin{tikzpicture}
    \begin{groupplot}[
        group style={
            group name=myplots,
            group size=8 by 1,
            horizontal sep=0.03cm
        },
        ybar,
        scale only axis,
        title style={font=\scriptsize, align=center},
        ylabel style={font=\scriptsize, align=center},
        tick label style={font=\scriptsize},
        x tick label style={rotate=45, anchor=east, font=\scriptsize},
        ymajorgrids=true,
        every axis plot/.append style={bar width=4.0pt, bar shift=0pt},
        width=1.90cm,
        height=1.8cm,
    ]

    \nextgroupplot[
        title={OBJECT},
        title style={yshift=-0.2cm},
        ylabel={AUC (\%)},
        ymin=30, ymax=95,
        ytick={40, 50,60,70,80,90},
        xmin=-1, xmax=5,
        xtick={0,1,2,3,4},
        xticklabels={\textit{BM}, \textit{EMA}, \textit{OM}, \textit{OEMA}, \textit{Ours}},
        xtick pos=left,
        ytick pos=left,
    ]
        \addplot[
            fill=gray!50,
            draw=gray!70,
            error bars/.cd,
            y dir=both,
            y explicit
        ] coordinates {
            (0,84.5) +- (0.4,0.4)
            (1,85.1) +- (2.2,2.2)
            (2,86.1) +- (2.0,2.0)
            (3,85.0) +- (2.2,2.2)
        };
        \addplot[
            fill=babyblue,
            draw=black,
            error bars/.cd,
            y dir=both,
            y explicit
        ] coordinates {
            (4,91.5) +- (0.1,0.1)
        };

    \nextgroupplot[
        title={SPATIAL},
        title style={yshift=-0.2cm},
        ymin=30, ymax=95,
        ytick={40, 50,60,70,80,90},
        yticklabels={},
        xmin=-1, xmax=5,
        xtick={0,1,2,3,4},
        xticklabels={\textit{BM}, \textit{EMA}, \textit{OM}, \textit{OEMA}, \textit{Ours}},
        xtick pos=left,
        ytick pos=left,
    ]
        \addplot[
            fill=gray!50,
            draw=gray!70,
            error bars/.cd,
            y dir=both,
            y explicit
        ] coordinates {
            (0,51.3) +- (1.2,1.2)
            (1,52.8) +- (1.9,1.9)
            (2,54.0) +- (1.3,1.3)
            (3,51.3) +- (2.1,2.1)
        };
        \addplot[
            fill=babyblue,
            draw=black,
            error bars/.cd,
            y dir=both,
            y explicit
        ] coordinates {
            (4,58.7) +- (1.0,1.0)
        };

    \nextgroupplot[
        hide axis,
        scale only axis,
        width=0.35cm,
        height=1.8cm,
        xtick=\empty,
        ytick=\empty,
        xlabel={},
        ylabel={},
        title={}
    ]

    \nextgroupplot[
        title={OBJECT},
        title style={yshift=-0.2cm},
        ymin=30, ymax=95,
        ytick={40, 50,60,70,80,90},
        yticklabels={},
        xmin=-1, xmax=3,
        xtick={0,1,2},
        xticklabels={\textit{Hinge}, \textit{InfoNCE}, \textit{Ours}},
        xtick pos=left,
        ytick pos=left,
    ]
        \addplot[
            fill=gray!50,
            draw=gray!70,
            error bars/.cd,
            y dir=both,
            y explicit
        ] coordinates {
            (0,85.5) +- (2.4,2.4)
            (1,87.5) +- (2.1,2.1)
        };
        \addplot[
            fill=babyblue,
            draw=black,
            error bars/.cd,
            y dir=both,
            y explicit
        ] coordinates {
            (2,91.5) +- (0.1,0.1)
        };

    \nextgroupplot[
        title={SPATIAL},
        title style={yshift=-0.2cm},
        ymin=30, ymax=95,
        ytick={40, 50,60,70,80,90},
        yticklabels={},
        xmin=-1, xmax=3,
        xtick={0,1,2},
        xticklabels={\textit{Hinge}, \textit{InfoNCE}, \textit{Ours}},
        xtick pos=left,
        ytick pos=left,
    ]
        \addplot[
            fill=gray!50,
            draw=gray!70,
            error bars/.cd,
            y dir=both,
            y explicit
        ] coordinates {
            (0,53.5) +- (1.9,1.9)
            (1,56.7) +- (2.5,2.5)
        };
        \addplot[
            fill=babyblue,
            draw=black,
            error bars/.cd,
            y dir=both,
            y explicit
        ] coordinates {
            (2,58.7) +- (1.0,1.0)
        };

    \nextgroupplot[
        hide axis,
        scale only axis,
        width=0.35cm,
        height=1.8cm,
        xtick=\empty,
        ytick=\empty,
        xlabel={},
        ylabel={},
        title={}
    ]

    \nextgroupplot[
        title={OBJECT},
        title style={yshift=-0.2cm},
        ymin=30, ymax=95,
        ytick={40, 50,60,70,80,90},
        yticklabels={},
        xmin=-1, xmax=4,
        xtick={0,1,2,3},
        xticklabels={\textit{MS-Rank}, \textit{MS-Sum}, \textit{Ctx-only}, \textit{Ours}},
        xtick pos=left,
        ytick pos=left,
    ]
        \addplot[
            fill=gray!50,
            draw=gray!70,
            error bars/.cd,
            y dir=both,
            y explicit
        ] coordinates {
            (0,88.3) +- (0.5,0.5)
            (1,85.6) +- (2.0,2.0)
            (2,74.0) +- (1.2,1.2)
        };
        \addplot[
            fill=babyblue,
            draw=black,
            error bars/.cd,
            y dir=both,
            y explicit
        ] coordinates {
            (3,91.5) +- (0.1,0.1)
        };

    \nextgroupplot[
        title={SPATIAL},
        title style={yshift=-0.2cm},
        ymin=30, ymax=95,
        ytick={40, 50,60,70,80,90},
        yticklabels={},
        xmin=-1, xmax=4,
        xtick={0,1,2,3},
        xticklabels={\textit{MS-Rank}, \textit{MS-Sum}, \textit{Ctx-only}, \textit{Ours}},
        xtick pos=left,
        ytick pos=left,
    ]
        \addplot[
            fill=gray!50,
            draw=gray!70,
            error bars/.cd,
            y dir=both,
            y explicit
        ] coordinates {
            (0,55.0) +- (3.0,3.0)
            (1,51.5) +- (4.0,4.0)
            (2,36.9) +- (3.4,3.4)
        };
        \addplot[
            fill=babyblue,
            draw=black,
            error bars/.cd,
            y dir=both,
            y explicit
        ] coordinates {
            (3,58.7) +- (1.0,1.0)
        };

    \end{groupplot}

    \node[font=\bfseries\scriptsize, anchor=south]
        at ($(myplots c1r1.north)!0.5!(myplots c2r1.north)+(0,0.3cm)$) {Prototype};

    \node[font=\bfseries\scriptsize, anchor=south]
        at ($(myplots c4r1.north)!0.5!(myplots c5r1.north)+(0,0.3cm)$) {Inter-task};

    \node[font=\bfseries\scriptsize, anchor=south]
        at ($(myplots c7r1.north)!0.5!(myplots c8r1.north)+(0,0.3cm)$) {Routing};

    \end{tikzpicture}
    \caption{AUC comparison of prototype estimation (left), inter-task regularization (center), and routing strategies (right) on LIBERO-OBJECT and LIBERO-SPATIAL, respectively. Gray bars denote baseline variants, and blue bars denote our method. See the text for abbreviation definitions.}
    \label{fig:ablation_auc_1x6}
\end{figure*}
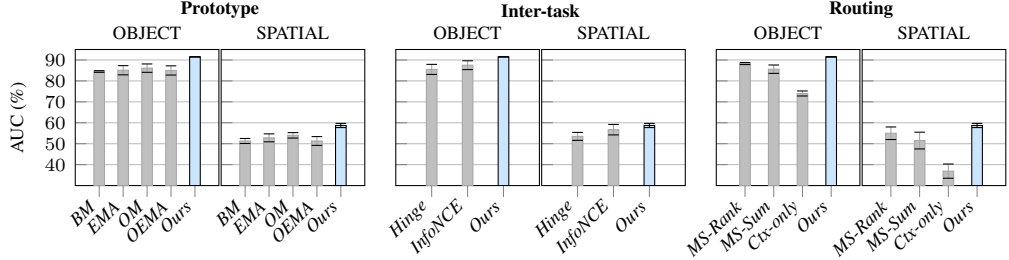

\textbf{Comparison of prototype estimation strategies in intra-task regularization.}
We compare different strategies for computing the task prototype used in intra-task regularization (Eq. \ref{eq:prototype}) in Fig.~\ref{fig:ablation_auc_1x6}-left. The \emph{Batch Mean} (BM) computes the prototype as the average of context representations within each mini-batch, without maintaining a persistent estimate across the task. The \emph{EMA} updates the prototype using an exponential moving average over observed representations. We also include oracle variants that assume access to the full dataset: the \emph{Oracle Mean} (OM), which computes the prototype using all samples, and the \emph{Oracle EMA} (OEMA), which applies an exponential moving average over the full dataset.
In contrast, our method maintains a \emph{Running Mean} (RM) of context representations throughout the task (Eq. \ref{eq:prototype}), 
which achieves the highest AUC across both benchmarks. Full results (including FWT and NBT), reported in Appendix Tab.~\ref{tab:key}, show that maintaining a consistent estimate over the task outperforms batch-level and momentum-based alternatives in AUC, with trade-offs in FWT and NBT.

\textbf{Comparison of inter-task regularization strategies.}
We compare different strategies for enforcing separation between task-specific routing keys (Fig.~\ref{fig:ablation_auc_1x6}-middle). In addition to our \emph{orthogonality-based} objective, we consider a \emph{hinge loss} that enforces a margin between pairs of context keys~\cite{huang2024class}, and an \emph{InfoNCE-style} objective that promotes separation via contrastive learning~\cite{song2023infocl}.
Our orthogonality loss enforces geometric separation by encouraging pairwise orthogonality between context keys, leading to a more structured routing space and improved AUC across benchmarks. Appendix Tab.~\ref{tab:inter} shows that these gains are accompanied by stable FWT and NBT.

\textbf{Effect of different routing strategies.}
Fig.~\ref{fig:ablation_auc_1x6}-right compares different routing strategies for combining instruction- and context-level signals. All methods use the same attention scores $a_{j,t}^{\text{ins}}$ and $a_{j,t}^{\text{ctx}}$, differing only in how these signals are combined and constrained.
The \emph{mixed score ranking} (MS-Rank) strategy treats instruction and context scores independently by pooling them together and selecting the top-$K$ values. In contrast, \emph{mixed score sum} (MS-Sum) first combines the two scores for each expert and then selects experts based on the aggregated values.
The \emph{no mask} (Ctx-only) variant removes the restriction to set $\mathcal{S}_{j^*}$, allowing selection from all experts.
Our method performs hierarchical routing as described in Sec.~\ref{sec:method}.
Our approach achieves the best AUC across benchmarks. Full results (including FWT and NBT) in Appendix Tab.~\ref{tab:router} show that removing masking significantly degrades performance, highlighting the importance of restricting routing to pre-selected experts. While heuristic combinations (e.g., MS-Rank) can reduce forgetting, our method achieves stronger FWT and overall performance. Overall, structured routing, combining hierarchical selection, masking, and learned fusion, is key to improved performance.

\subsubsection{Expert Activation Patterns}
\begin{figure}[tb!]
    \centering
    
    \begin{subfigure}[b]{0.24\columnwidth}
        \centering
        \includegraphics[width=\linewidth]{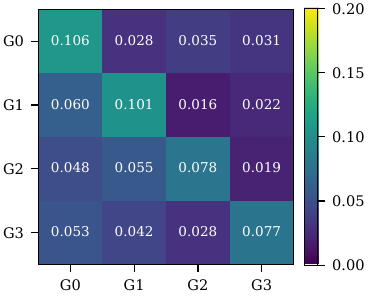}
        {\scriptsize \caption{DMPEL}}
        \label{fig:attn_dmpel}
    \end{subfigure}
    \hfill
    \begin{subfigure}[b]{0.24\columnwidth}
        \centering
        \includegraphics[width=\linewidth]{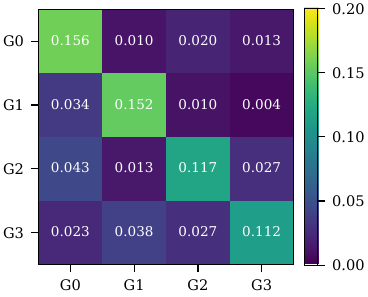}
        {\scriptsize \caption{$\mathrm{DualAttn}(a^{\text{ins}}_{j,t}, a^{\text{ctx}}_{j,t})$}}
        \label{fig:attn_dual}
    \end{subfigure}
    \hfill
    \begin{subfigure}[b]{0.24\columnwidth}
        \centering
        \includegraphics[width=\linewidth]{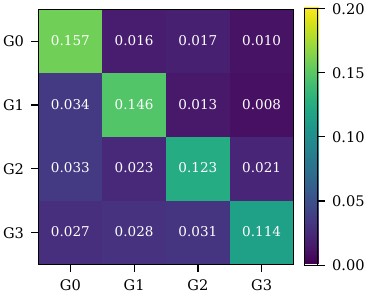}
        {\scriptsize \caption{$+\ \mathcal{L}_{\text{intra}}$}}
        \label{fig:attn_intra}
    \end{subfigure}
    \hfill
    \begin{subfigure}[b]{0.24\columnwidth}
        \centering
        \includegraphics[width=\linewidth]{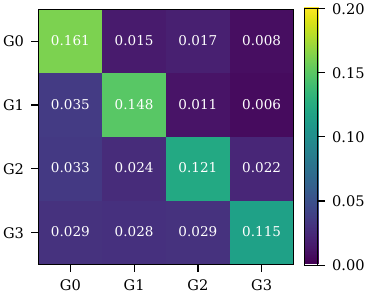}
        {\scriptsize \caption{$+\ \mathcal{L}_{\text{intra}} + \mathcal{L}_{\text{inter}}$ (Ours)}}
        \label{fig:attn_final}
    \end{subfigure}

    \caption{Expert activation heatmaps aggregated at the task-group level under different settings. Each matrix shows the average attention from query task groups (rows) to expert groups (columns). Diagonal entries indicate within-group routing, while off-diagonal entries reflect cross-group activation. }
    \label{fig:group4x4_compare}
\end{figure}

We analyze routing behavior on LIBERO-MIX, where tasks are drawn from four benchmark splits (OBJECT, SPATIAL, GOAL, and LONG), which we treat as task groups. As shown in Appendix Fig.~\ref{fig:query_vis}, context query embeddings from tasks within the same group are more similar than those across groups, indicating that the learned representations preserve task-level semantics.
To study routing behavior, we associate each expert with its corresponding task and assign it the same group label. During evaluation, we record the selected experts for each task and compute group-level routing statistics by aggregating attention weights across all tasks. Fig.~\ref{fig:group4x4_compare} visualizes expert activation patterns aggregated at the group level. Each heatmap shows the average attention from query task groups (rows) to expert groups (columns), where diagonal entries indicate within-group routing and off-diagonal entries reflect cross-group activation. A stronger diagonal indicates more semantically consistent routing. Compared to DMPEL, \textsc{hidra} exhibits a clearer concentration of attention along the diagonal, indicating a stronger preference for semantically aligned experts. Introducing $\mathcal{L}_{\text{intra}}$ further sharpens this behavior, while adding $\mathcal{L}_{\text{inter}}$ yields the most pronounced diagonal structure. These results demonstrate that the proposed regularization improves the consistency of expert selection and reduces cross-task interference.

\section{Conclusion}
\label{sec:conc}

We study LIL in a replay-free, single-pass setting where agents continuously adapt from multimodal demonstration streams without revisiting past data. We identify routing interference as a key challenge in modular architectures, degrading expert selection over time. Consequently, we propose a structured routing framework with dual attention that separates instruction retrieval from context refinement. Combined with key-level geometric regularization and expandable experts, it enables stable expert selection during sequential adaptation without replay. Experiments on LIBERO show consistent gains in forward transfer and overall performance with low forgetting, outperforming all SOTA. \\
\textbf{Limitations and Future Work.} Performance is strong overall, but drops on LIBERO-SPATIAL, where fine-grained spatial differences require precise context over instruction semantics, making routing sensitive to subtle perceptual variations. Improving spatial representations, context modeling, and evaluation on real robotic platforms are key future directions. \\
\textbf{Broader Impacts.} The proposed approach supports more adaptive and efficient robotic systems for real-world applications. As with other learning-based methods, deployment in safety-critical settings may introduce risks due to perception or decision errors, requiring careful evaluation and robust deployment practices.

\bibliographystyle{unsrt}
\bibliography{refs}

\clearpage

\appendix
 
\section{Technical Appendices and Supplementary Material}

The supplementary material provides extended analyses and implementation details that complement the main paper. Specifically, it includes: (i) extended ablation studies covering the effects of regularization weights, number of selected experts, routing projection capacity, and number of demonstrations per task; (ii) additional evaluation of routing behavior, including quantitative statistics and selection patterns; (iii) computational and storage efficiency analysis, including FLOPs, inference time, and per-task memory overhead; (iv) results under multi-pass training to assess robustness beyond the single-pass setting; and (v) full task and instruction lists for the LIBERO-MIX and LIBERO-INS benchmarks to ensure reproducibility. These materials provide deeper insight into the proposed method’s behavior, design choices, and empirical performance. For clarity and consistency, the final model configuration is highlighted in all tables using a \textcolor{babyblue}{\rule{0.8em}{0.8em}} marker.


\subsection{Implementation Details}
\label{sec:imple}
\textbf{Architecture and Multi-modal Fusion.} Following established LIL protocols \cite{lei2025dynamic}, we employ pre-trained CLIP encoders to extract visual and language representations. Proprioceptive states are encoded through a single-layer MLP. To integrate these modalities, we use Feature-wise Linear Modulation (FiLM) \cite{perez2018film}, where the language features condition both the visual and state representations to produce task-aware embeddings. These fused features are then processed by a temporal decoder based on the GPT-2 architecture \cite{achiam2023gpt} over a sliding temporal window. The resulting latent representation is mapped to the action space by a dedicated policy head. During the pre-training phase, all backbone components are optimized jointly.

\textbf{Expert LoRA Configurations.} We attach LoRA experts to the linear layers of the frozen encoders and the temporal backbone. Following \cite{liu2023tail,lei2025dynamic}, we use a fixed rank $r=16$ for all LoRA experts. This low-rank constraint ensures that the parameter growth per task remains minimal while providing sufficient capacity for task-specific adaptation.

\textbf{Routing Queries and Keys.} The \textsc{hidra} routing mechanism relies on a dual-key space with distinct initialization strategies. For the instruction branch, the keys $k_j^{\mathrm{ins}}$ are initialized directly from the language embeddings of the corresponding task instructions. This provides a semantically meaningful starting point for task-level retrieval, ensuring that the Stage I selection is anchored in the task’s linguistic definition. In contrast, the context keys $k_j^{\mathrm{ctx}}$ are initialized randomly and optimized during the training phase to capture fine-grained environmental cues, governed by the key-space regularization losses. At inference, the instruction query $q_t^{\mathrm{ins}}$ is derived from the current language embedding, while the context query $q_t^{\mathrm{ctx}}$ is projected from the fused multimodal state $r_t$ to facilitate state-dependent expert refinement.

\textbf{Task-specific Routing Projection.}
The routing projection $\phi_j(\cdot)$ is implemented as a lightweight task-specific MLP.
For each task-specific routing branch, we first use two separate one-layer projection networks to encode the instruction-level and context-level routing evidence:
\begin{equation}
\mathbf{u}^{\mathrm{ins}}_{j,t}
=
f^{\mathrm{ins}}_j
\left(
a^{\mathrm{ins}}_{j,t},
\mathbf{q}^{\mathrm{ins}}_{t} \odot \mathbf{k}^{\mathrm{ins}}_{j}
\right)
\in \mathbb{R}^{d_u},
\qquad
\mathbf{u}^{\mathrm{ctx}}_{j,t}
=
f^{\mathrm{ctx}}_j
\left(
a^{\mathrm{ctx}}_{j,t},
\mathbf{q}^{\mathrm{ctx}}_{t} \odot \mathbf{k}^{\mathrm{ctx}}_{j}
\right)
\in \mathbb{R}^{d_u}.
\end{equation}

The resulting hidden features are concatenated and passed through another one-layer MLP network followed by a sigmoid activation to obtain the final fused routing score:
\begin{equation}
s^{\mathrm{fused}}_{j,t}
=
\sigma
\left(
g_j
\left(
\left[
\mathbf{u}^{\mathrm{ins}}_{j,t};
\mathbf{u}^{\mathrm{ctx}}_{j,t}
\right]
\right)
\right),
\end{equation}
where $\sigma(\cdot)$ is the sigmoid function.
In our implementation, we set the hidden dimension to $d_u=8$.
This task-specific projection enables the router to adaptively combine instruction-level and context-level evidence for each task, while introducing only a small amount of additional computation.

\subsection{Extended Ablation Study}

\textbf{Effect of regularization weights.}
We analyze the impact of the weighting coefficients $\lambda_{\text{intra}}$ and $\lambda_{\text{inter}}$ in the overall objective, as shown in Tab.~\ref{tab:lambda}. These coefficients control the relative importance of intra-task alignment and inter-task separation in shaping the routing space. We set $\lambda_{\text{intra}} = \lambda_{\text{inter}} = 1$ as the default configuration, corresponding to equal coefficient weighting between the behavioral cloning objective and the regularization terms. Starting from this setting, we vary the coefficients by decreasing their values to study the effect of weaker alignment or separation constraints. Increasing the weights beyond this point would primarily amplify the contribution of the regularization terms, without providing additional insight into their relative roles.

As shown in Tab.~\ref{tab:lambda}, the choice of $\lambda_{\text{intra}}$ and $\lambda_{\text{inter}}$ significantly affects the trade-off between performance and stability. When $\lambda_{\text{inter}}$ is small (e.g., $0.01$), the model achieves the highest forward transfer (FWT), but suffers from severe forgetting, as reflected by large NBT values. Conversely, intermediate values of $\lambda_{\text{inter}}$ (e.g., $0.1$) improve stability, yielding the lowest NBT across benchmarks, but at the cost of reduced overall performance.
Reducing $\lambda_{\text{intra}}$ leads to consistent degradation in both FWT and AUC, indicating that intra-task alignment is critical for maintaining strong performance. In contrast, varying $\lambda_{\text{inter}}$ primarily affects the level of forgetting, suggesting that inter-task regularization plays a key role in controlling interference.
The configuration $\lambda_{\text{intra}} = 1.0$ and $\lambda_{\text{inter}} = 1.0$ provides the best overall balance, achieving the highest AUC across both benchmarks while maintaining competitive forward transfer and moderate forgetting. These results highlight a clear trade-off between adaptation and stability, and show that jointly enforcing alignment and separation is necessary for optimal performance.

\begin{table}[b!]
\centering
\caption{
The impact of the weighting coefficients $\lambda_{\text{intra}}$ and $\lambda_{\text{inter}}$.}
\label{tab:lambda}
\resizebox{0.7\linewidth}{!}{
\begin{tabular}{cc|ccc|ccc}
\toprule
\multirow{2}{*}{$\lambda_{\text{intra}}$} & \multirow{2}{*}{$\lambda_{\text{inter}}$}
& \multicolumn{3}{c|}{\textbf{LIBERO-OBJECT}}
& \multicolumn{3}{c}{\textbf{LIBERO-SPATIAL}} \\
\cmidrule(lr){3-5} \cmidrule(lr){6-8}
 & 
 & FWT$\uparrow$ & NBT$\downarrow$ & AUC$\uparrow$
 & FWT$\uparrow$ & NBT$\downarrow$ & AUC$\uparrow$ \\
\midrule

1.0 & 0.01  & \textbf{93.5 $\pm$1.2} & 6.4 $\pm$1.7 & 89.0 $\pm$0.5  & 
56.0$\pm$4.0 & 0.4$\pm$4.5 & 56.3$\pm$1.4
\\
1.0 & 0.1 & 84.0 $\pm$2.1 & \textbf{-2.4 $\pm$1.1} & 85.6 $\pm$0.9  & 51.0 $\pm$1.8 & \textbf{-4.8 $\pm$2.3} & 54.4 $\pm$1.7  \\
0.01 & 1.0 &  87.0 $\pm$0.7 & -1.3 $\pm$1.2 & 87.4 $\pm$0.6  & 52.0 $\pm$1.3 & -0.4 $\pm$2.5 & 52.4 $\pm$2.7  \\ 
0.1 & 1.0  & 86.5 $\pm$2.1 & -0.1 $\pm$1.2 & 86.6 $\pm$1.8  & 54.5 $\pm$1.6 & 3.0 $\pm$3.2 & 52.8 $\pm$2.1  \\
\rowcolor{babyblue}
1.0 & 1.0  & \textbf{93.5$\pm$1.5} & 2.2$\pm$1.9 & \textbf{91.5$\pm$0.1}
   & 
   \textbf{58.0 $\pm$1.5} & -1.4 $\pm$0.8 & \textbf{58.7 $\pm$1.0}
   \\  
\bottomrule
\end{tabular}}
\end{table}


\textbf{Effect of the number of selected experts.}
Tab.~\ref{tab:ablation_expert} evaluates the impact of the number of selected experts $K$ in the Top-$K$ routing mechanism. 
In our hierarchical routing design, $K$ controls how many experts are combined.
We observe that performance is sensitive to $K$, with $K=3$ achieving the best results in terms of AUC and forward transfer across both benchmarks. In contrast, smaller values (e.g., $K=2$) yield lower forgetting, indicating a trade-off between selective routing and stability. Larger values (e.g., $K=4$) do not provide further gains and can slightly degrade performance.
Overall, these results suggest that a moderate number of selected experts provides the best balance between performance and stability, while the method remains robust across different choices of $K$.

\begin{table}[tb!]
\centering
\caption{Influence of the number of selected experts on performance.}
\label{tab:ablation_expert}
\resizebox{0.7\linewidth}{!}{
\begin{tabular}{lcccccc}
\toprule
\textbf{\# Experts} & \multicolumn{3}{c}{\textbf{LIBERO-OBJECT}} & \multicolumn{3}{c}{\textbf{LIBERO-SPATIAL}} \\
\cmidrule(lr){2-4} \cmidrule(lr){5-7}
 & FWT$\uparrow$ & NBT$\downarrow$ & AUC$\uparrow$ & FWT$\uparrow$ & NBT$\downarrow$ & AUC$\uparrow$ \\
\midrule
2 & 89.0$\pm$1.6 & \textbf{0.1$\pm$1.6} & 88.6$\pm$2.2 & 53.0$\pm$1.9 & \textbf{-2.6$\pm$1.4} & 54.7$\pm$2.0 \\
\rowcolor{babyblue}
3 & \textbf{93.0$\pm$1.5} & 2.2$\pm$1.9 & \textbf{91.5$\pm$0.1}
    & \textbf{58.0 $\pm$1.5} & -1.4 $\pm$0.8 & \textbf{58.7 $\pm$1.0}  \\
4 & 87.0$\pm$1.5 & 2.8$\pm$1.4 & 84.9$\pm$1.7 & 56.5$\pm$1.7 & -0.2$\pm$1.2 & 56.3$\pm$1.8 \\
\bottomrule
\end{tabular}}
\end{table}

\textbf{Effect of the task-specific routing projection dimension $d_u$.}
Tab.~\ref{tab:ablation_mlp} evaluates the impact of $d_u$, the feature dimension of $\mathbf{u}^{\mathrm{ins}}_{j,t}$ and $\mathbf{u}^{\mathrm{ctx}}_{j,t}$ produced by the task-specific routing projection networks $f^{\mathrm{ins}}_j(\cdot)$ and $f^{\mathrm{ctx}}_j(\cdot)$ in Sec.~\ref{sec:imple}.
We observe a non-monotonic trend, where moderate capacity yields the best trade-off between performance and stability. In particular, a hidden dimension of 8 achieves the highest AUC on LIBERO-OBJECT and strong performance across all metrics, and is therefore used in the final model. Smaller dimensions (e.g., 4) lead to reduced performance, suggesting insufficient capacity to effectively fuse instruction and context signals. Larger dimensions (e.g., 16 and 32) do not provide consistent improvements. While they remain competitive and can improve specific metrics (e.g., NBT or FWT), they do not yield better overall performance. These results indicate that the projection module requires sufficient but controlled capacity to reliably combine routing signals, highlighting the importance of balanced model complexity in replay-free settings.

\begin{table}[tb!]
\centering
\caption{Influence of the hidden dimension of the MLP used to compute routing scores. The MLP has a fixed depth of two layers.}
\label{tab:ablation_mlp}
\resizebox{0.7\linewidth}{!}{
\begin{tabular}{lcccccc}
\toprule
\textbf{MLP Dim.} & \multicolumn{3}{c}{\textbf{LIBERO-OBJECT}} & \multicolumn{3}{c}{\textbf{LIBERO-SPATIAL}} \\
\cmidrule(lr){2-4} \cmidrule(lr){5-7}
 & FWT$\uparrow$ & NBT$\downarrow$ & AUC$\uparrow$ & FWT$\uparrow$ & NBT$\downarrow$ & AUC$\uparrow$ \\
\midrule
4  & 86.5$\pm$2.4 & 1.5$\pm$2.4 & 85.2$\pm$2.0 & 47.0$\pm$2.6 & \textbf{-5.4$\pm$2.1} & 50.9$\pm$1.6 \\
\rowcolor{babyblue}
8  & \textbf{93.0$\pm$1.5} & 2.2$\pm$1.9 & \textbf{91.5$\pm$0.1}
   & \textbf{58.0 $\pm$1.5} & -1.4 $\pm$0.8 & \textbf{58.7 $\pm$1.0}  \\
16 & 83.0$\pm$1.4 & 1.2$\pm$2.0 & 82.6$\pm$1.6 & 56.0$\pm$1.2 & 5.0$\pm$1.3 & 53.0$\pm$1.4 \\
32 & 83.0$\pm$1.2 & \textbf{-3.2$\pm$1.5} & 85.6$\pm$1.2 & \textbf{58.0$\pm$1.6} & 3.0$\pm$3.0 & 56.8$\pm$1.8 \\
\bottomrule
\end{tabular}}
\end{table}

\textbf{Effect of the number of demonstrations per task.} As expected, increasing the number of demonstrations improves performance in terms of FWT and AUC across both benchmarks (Tab.~\ref{tab:num_demo}). The largest gains occur when moving from 10 to 20 demonstrations, highlighting the benefit of additional supervision in low-data regimes.
The results also demonstrate the data efficiency of our \textsc{HIDRA}. With only 20 demonstrations per task, the model already achieves strong performance (e.g., AUC of 71.2 on LIBERO-OBJECT). Increasing to 50 demonstrations further improves performance, yielding the best overall results and showing that the method scales effectively with more data.
In contrast, NBT does not follow a monotonic trend. While forgetting remains generally well-controlled across all settings, it does not consistently improve with more data. This suggests that the proposed routing mechanism enables both effective learning from additional demonstrations and stable retention of previously acquired knowledge.

\begin{table}[tb!]
\centering
\caption{Effect of the number of demonstrations per task.}
\label{tab:num_demo}
\resizebox{0.7\linewidth}{!}{
\begin{tabular}{lccccccccc}
\toprule
\textbf{} & \multicolumn{3}{c}{\textbf{LIBERO-OBJECT}} & \multicolumn{3}{c}{\textbf{LIBERO-SPATIAL}} \\
\cmidrule(lr){2-4} \cmidrule(lr){5-7}
 & FWT$\uparrow$ & NBT$\downarrow$ & AUC$\uparrow$ & FWT$\uparrow$ & NBT$\downarrow$ & AUC$\uparrow$ \\
\midrule
10   & 54.0$\pm$2.2 & 5.9$\pm$1.7 & 49.4$\pm$0.8
   & 16.5$\pm$1.9 & \textbf{-2.7$\pm$3.0} & 18.6$\pm$3.5 \\
20
   & 70.0$\pm$4.6 & \textbf{-2.1$\pm$3.6} & 71.2$\pm$2.4  & 40.0$\pm$1.6 & -0.8$\pm$1.2 & 40.4$\pm$1.6 \\
\rowcolor{babyblue}
50  & \textbf{93.0$\pm$1.5} & 2.2$\pm$1.9 & \textbf{91.5$\pm$0.1}
     & \textbf{58.0 $\pm$1.5} & -1.4 $\pm$0.8 & \textbf{58.7 $\pm$1.0}  \\
\bottomrule

\end{tabular}}
\end{table}



\textbf{Prototype estimation strategies.}
Tab.~\ref{tab:key} reports the full results, in terms of FWT, NBT, and AUC, corresponding to Fig.~\ref{fig:ablation_auc_1x6}-left in the main paper.
As shown in Tab.~\ref{tab:key}, our approach achieves the best performance in terms of AUC and FWT across benchmarks. While EMA-based methods yield lower NBT, the running mean provides a better overall trade-off between performance and stability. These results suggest that maintaining a consistent estimate over the task is beneficial compared to batch-level or momentum-based alternatives.

\begin{table}[tb!]
\centering
\caption{Comparison of prototype estimation strategies for intra-task regularization.}
\label{tab:key}
\resizebox{0.8\linewidth}{!}{
\begin{tabular}{lcccccc}
\toprule
\textbf{} & \multicolumn{3}{c}{\textbf{LIBERO-OBJECT}} & \multicolumn{3}{c}{\textbf{LIBERO-SPATIAL}} \\
\cmidrule(lr){2-4} \cmidrule(lr){5-7}
 & FWT$\uparrow$ & NBT$\downarrow$ & AUC$\uparrow$ 
 & FWT$\uparrow$ & NBT$\downarrow$ & AUC$\uparrow$ \\
\midrule

Batch Mean 
& 88.0$\pm$1.9 & 2.5$\pm$1.4 & 84.5$\pm$0.4
& 54.0$\pm$2.0 & 1.3$\pm$2.5 & 51.3$\pm$1.2 \\

EMA 
& 81.5$\pm$1.8 & \textbf{-4.3$\pm$1.5} & 85.1$\pm$2.2
& 49.0$\pm$2.6 & \textbf{-6.3$\pm$2.4} & 52.8$\pm$1.9 \\

Oracle Mean 
& 87.0$\pm$1.6 & 0.3$\pm$1.8 & 86.1$\pm$2.0
& 49.5$\pm$2.5 & -4.0$\pm$2.5 & 54.0$\pm$1.3 \\

Oracle EMA 
& 82.0$\pm$3.6 & -2.3$\pm$0.8 & 85.0$\pm$2.2
& 53.0$\pm$1.6 & 2.2$\pm$2.5 & 51.3$\pm$2.1 \\

\rowcolor{babyblue}
Running Mean (Ours)
& \textbf{93.0$\pm$1.5} & 2.2$\pm$1.9 & \textbf{91.5$\pm$0.1}
& \textbf{58.0 $\pm$1.5} & -1.4 $\pm$0.8 & \textbf{58.7$\pm$1.0}  \\

\bottomrule
\end{tabular}}
\end{table}


\textbf{Inter-task regularization strategies.}
Tab.~\ref{tab:inter} reports the full results in terms of FWT, NBT, and AUC, corresponding to Fig.~\ref{fig:ablation_auc_1x6}-center. As shown in Tab.~\ref{tab:inter}, it achieves the best performance in terms of AUC and FWT across benchmarks. While alternative objectives (e.g., InfoNCE) yield lower NBT, the orthogonality loss provides a better overall trade-off between performance and stability. These results suggest that explicitly imposing geometric structure in the routing space improves overall task performance, albeit with a trade-off in stability.

\begin{table}[tb!]
\centering
\caption{Comparison of inter-task regularization strategies.}
\label{tab:inter}
\resizebox{0.8\linewidth}{!}{
\begin{tabular}{lccccccccc}
\toprule
& \multicolumn{3}{c}{\textbf{LIBERO-OBJECT}} & \multicolumn{3}{c}{\textbf{LIBERO-SPATIAL}} \\
\cmidrule(lr){2-4} \cmidrule(lr){5-7}
 & FWT$\uparrow$ & NBT$\downarrow$ & AUC$\uparrow$ & FWT$\uparrow$ & NBT$\downarrow$ & AUC$\uparrow$ \\
\midrule
Hinge loss  & 82.0 $\pm$2.6 &  -2.6$\pm$1.4 &  85.5$\pm$ 2.4 
& 56.5$\pm$2.2 &  5.0$\pm$2.9 & 53.5$\pm$ 1.9 
\\
InfoNCE loss & 85.0 $\pm$1.7 &  \textbf{-3.6$\pm$0.9} &  87.5$\pm$ 2.1 
& 55.0$\pm$1.8 &  \textbf{-2.6$\pm$2.5} & 56.7$\pm$ 2.5 
\\
\rowcolor{babyblue}
Orthogonality loss (Ours) & \textbf{93.0$\pm$1.5} & 2.2$\pm$1.9 & \textbf{91.5$\pm$0.1}
    & \textbf{58.0 $\pm$1.5} & -1.4 $\pm$0.8 & \textbf{58.7 $\pm$1.0}  \\
\bottomrule

\end{tabular}}
\end{table}


\textbf{Routing strategies.}
Tab.~\ref{tab:router} reports the full results in terms of FWT, NBT, and AUC, corresponding to Fig.~\ref{fig:ablation_auc_1x6}-rigth.
As seen, our approach achieves the best performance in terms of AUC and FWT. Removing the masking mechanism leads to a significant degradation, highlighting the importance of restricting routing to pre-selected experts. While heuristic combinations (e.g., mixed ranking) can yield lower forgetting, learned fusion provides a better overall trade-off between performance and stability. Overall, structured routing, which combines hierarchical selection, masking, and learned fusion, is key to improved performance.

\begin{table}[tb!]
\centering
\caption{Comparison of different routing strategies.}
\label{tab:router}
\resizebox{0.8\linewidth}{!}{
\begin{tabular}{lcccccc}
\toprule
 & \multicolumn{3}{c}{\textbf{LIBERO-OBJECT}} & \multicolumn{3}{c}{\textbf{LIBERO-SPATIAL}} \\
\cmidrule(lr){2-4} \cmidrule(lr){5-7}
 & FWT$\uparrow$ & NBT$\downarrow$ & AUC$\uparrow$ 
 & FWT$\uparrow$ & NBT$\downarrow$ & AUC$\uparrow$ \\
\midrule

Mixed Score Ranking 
& 91.5 $\pm$ 1.0 & \textbf{-3.6 $\pm$ 0.8} & 88.3 $\pm$ 0.5
& 52.5 $\pm$ 3.6 & \textbf{-2.0 $\pm$ 2.9} & 55.0 $\pm$ 3.0 \\

Mixed Score Sum 
& 87.5 $\pm$ 2.2 & 3.1 $\pm$ 1.8 & 85.6 $\pm$ 2.0
& 50.0 $\pm$ 3.4 & -1.2 $\pm$ 2.8 & 51.5 $\pm$ 4.0 \\

Context-only (No Mask) 
& 89.5 $\pm$ 1.2 & 19.1 $\pm$ 1.5 & 74.0 $\pm$ 1.2
& 55.0 $\pm$ 2.4 & 23.1 $\pm$ 3.8 & 36.9 $\pm$ 3.4 \\

\rowcolor{babyblue}
\textsc{hidra} (Ours)
& \textbf{93.0 $\pm$ 1.5} & 2.2 $\pm$ 1.9 & \textbf{91.5 $\pm$ 0.1}
  & \textbf{58.0 $\pm$1.5} & -1.4 $\pm$0.8 & \textbf{58.7 $\pm$1.0}  \\

\bottomrule
\end{tabular}}
\end{table}

\begin{table}[b!]
\centering
\caption{Routing statistics on the last four tasks of \textbf{LIBERO-MIX}. 
SG and CG denote same-group and cross-group proportions of selected experts and attention mass, respectively. 
Entropy measures the dispersion of routing decisions. Higher SG values indicate more semantically consistent routing. See text for more details.}
\label{tab:router_stats_last4}
\resizebox{\columnwidth}{!}{
\begin{tabular}{lcccccc}
\toprule
\textbf{} & \textbf{SG-Count} & \textbf{SG-Attn} & \textbf{CG-Count} & \textbf{CG-Attn} & \textbf{Ent-Count} & \textbf{Ent-Attn}  \\
\midrule
$\text{DualAttn}(a^{\text{ins}}_{j,t}, a^{\text{ctx}}_{j,t})$ & $0.5829 \pm 0.1606$ & $0.6830 \pm 0.1192$ & $0.4171 \pm 0.1606$ & $0.3170 \pm 0.1192$ & $1.6799 \pm 0.0866$ & $1.4780 \pm 0.0803$  \\
 $+\ \mathcal{L}_{\text{intra}}$   & $0.6253 \pm 0.1400$ & $0.7185 \pm 0.0987$ & $0.3747 \pm 0.1400$ & $0.2815 \pm 0.0987$ & $2.0324 \pm 0.0546$ & $1.7496 \pm 0.0444$  \\
 \rowcolor{babyblue}
 $+\ \mathcal{L}_{\text{intra}} + \mathcal{L}_{\text{inter}}$  (Ours)      & $0.6466 \pm 0.1454$ & $0.7339 \pm 0.1040$ & $0.3534 \pm 0.1454$ & $0.2661 \pm 0.1040$ & $2.0376 \pm 0.0942$ & $1.7577 \pm 0.0787$  \\
DMPEL (100\%) & $0.3784 \pm 0.0879$ & $0.4254 \pm 0.1208$ & $0.6216 \pm 0.0879$ & $0.5746 \pm 0.1208$ & $2.3407 \pm 0.0908$ & $2.2335 \pm 0.0818$ \\
\bottomrule
\end{tabular}
}
\end{table}

\subsection{Router Selection Patterns}
\label{SUPPMATsec:routerSelect}

We analyze the routing behavior on the LIBERO-MIX, corresponding to the four benchmark splits (OBJECT, SPATIAL, GOAL, and LONG). Specifically, we measure how often the router selects experts from the same benchmark group as the query task (SG-Count), and how much attention mass is assigned to such experts (SG-Attn). We also report the corresponding cross-group statistics (CG-Count and CG-Attn), as well as the entropy of the selection and attention distributions (Ent-Count and Ent-Attn), which reflect how concentrated or dispersed the routing decisions are.

As shown in Tab.~\ref{tab:router_stats_last4}, the proposed regularization progressively makes routing more semantically consistent. Starting from the dual-attention baseline, adding $\mathcal{L}_{\text{intra}}$ increases both SG-Count and SG-Attn, while reducing CG-Count and CG-Attn, indicating that the router becomes more likely to select experts aligned with the query task. Introducing $\mathcal{L}_{\text{inter}}$ further strengthens this trend, yielding the highest same-group selection and attention among the proposed variants.
In contrast, DMPEL exhibits the opposite behavior, with significantly higher cross-group selection and attention, suggesting less semantically consistent routing. It also shows higher entropy, indicating more dispersed routing decisions compared to our method.
Overall, these results show that the proposed design not only structures the representation space, but also leads to more semantically consistent expert selection during routing.


\subsection{Computational and Storage Efficiency}
\label{SUPPMATsec:comp}

\begin{table}[tb!]
\centering
\caption{
Computational cost on \textbf{LIBERO-OBJECT}. All variants exhibit nearly identical forward latency and FLOPs, indicating that the proposed regularization and routing mechanisms introduce negligible computational overhead.
}
\small
\begin{tabular}{lccc}
\toprule
\textbf{} 
& \textbf{Forward Time (ms)} 
& \textbf{FLOPs (per forward)} 
& \textbf{Train FLOPs} \\
\midrule
Base (DualAttn)
& 125.2 $\pm$ 5.0 & $5.08\times10^{11}$ & $1.19\times10^{16}$ \\

+ $\mathcal{L}_{\mathrm{intra}}$
& 124.8 $\pm$ 3.8 & $5.08\times10^{11}$ & $1.19\times10^{16}$ \\

\rowcolor{babyblue}
+ $\mathcal{L}_{\mathrm{intra}} + \mathcal{L}_{\mathrm{inter}}$ (Ours)
& 125.8 $\pm$ 5.8 & $5.08\times10^{11}$ & $1.19\times10^{16}$ \\

\midrule
DMPEL (100\%)
& 125.5 $\pm$ 6.0 & $5.08\times10^{11}$ & $1.19\times10^{16}$  \\

\bottomrule
\end{tabular}
\label{tab:comp}
\end{table}

\begin{table}[tb!]
\centering
\caption{
Per-task memory overhead of our method. Storage is reported in KB and includes all task-specific components. Ratios indicate the contribution of each component to the total overhead. DMPEL is shown for comparison.
}
\small
\begin{tabular}{lcc}
\toprule
\textbf{Component} & \textbf{Storage (KB)} & \textbf{Ratio (\%)} \\
\midrule
Adapter-based experts & 2647 & 95.4 \\
Router parameters     & 125  & 4.5 \\
Auxiliary components  & 3    & 0.1 \\
\midrule
\rowcolor{babyblue}
Total (Ours)          & 2773 & 100 \\
\midrule
DMPEL (100\%)         & 48000 & 100 \\
\bottomrule
\end{tabular}
\label{tab:storage_breakdown}
\end{table}

We evaluate computational efficiency using three metrics, summarized in Tab.~\ref{tab:comp}. We consider a base DualAttn model and progressively add $\mathcal{L}{\mathrm{intra}}$ and $\mathcal{L}{\mathrm{inter}}$.
\textbf{Forward Time} measures the inference latency required to produce a single action (one forward pass). Results are obtained on an NVIDIA A100 GPU using 10 warm-up runs followed by 100 timed runs, reporting the mean and standard deviation.
\textbf{FLOPs (per forward)} denote the number of floating-point operations per forward pass, capturing the per-action inference cost. To approximate training cost, we report \textbf{Train-FLOPs}, computed as the total number of forward-pass FLOPs multiplied by the number of training iterations, with a factor of 3 accounting for forward, backward, and parameter update steps:
$
\text{Train-FLOPs} = \text{FLOPs} \times \text{training iterations} \times 3.
$

As shown in Tab.~\ref{tab:comp}, all variants exhibit nearly identical computational cost. Adding $\mathcal{L}{\mathrm{intra}}$ and $\mathcal{L}{\mathrm{inter}}$ does not modify the test-time architecture, resulting in negligible differences in Forward Time and FLOPs. Consequently, Train-FLOPs also remain unchanged. Compared to DMPEL, our method achieves similar computational efficiency without introducing additional overhead.

We further analyze per-task storage overhead in Tab.~\ref{tab:storage_breakdown}. As our approach introduces task-specific parameters without storing past data, only newly added components need to be retained. Under a fixed architecture and adapter configuration, each task requires approximately 2773 KB of additional storage. The majority of this cost arises from adapter-based experts, while the router contributes only a small fraction. In contrast, DMPEL incurs substantially higher storage overhead due to maintaining past data, requiring approximately $16\times$ more memory per task. These results highlight that our method significantly improves storage efficiency while avoiding the need for replay.

\subsection{Comparison with SOTA Methods under Multi-pass Training}
\label{secAppen:MultiPass}

\begin{table}[tb!]
\centering
\caption{Performance under multi-pass training (mean $\pm$ std). Best in \textbf{bold}.}
\label{tab:epoch}
\resizebox{0.8\linewidth}{!}{
\begin{tabular}{lcccccc}
\toprule
\multicolumn{7}{c}{\textbf{LIBERO-OBJECT}} \\
\midrule
\multirow{2}{*}{Method}
& \multicolumn{3}{c}{3-pass}
& \multicolumn{3}{c}{5-pass} \\
\cmidrule(lr){2-4}\cmidrule(lr){5-7}
& FWT$\uparrow$ & NBT$\downarrow$ & AUC$\uparrow$
& FWT$\uparrow$ & NBT$\downarrow$ & AUC$\uparrow$ \\
\midrule

MLR-IFA~\cite{yu2026lifelong}
& 85.7$\pm$1.6 & 2.5$\pm$0.3 & 84.3$\pm$0.7
& 80.0$\pm$1.3 & -3.1$\pm$2.7 & 83.6$\pm$1.4 \\

ISCIL~\cite{lee2024incremental}
& 85.5$\pm$1.4 & \textbf{-2.5$\pm$2.5} & 87.3$\pm$1.8
& 81.5$\pm$1.8 & \textbf{-7.6$\pm$1.9} & 86.9$\pm$1.2 \\

DMPEL (100\%)~\cite{lei2025dynamic}
& 81.0$\pm$1.5 & -2.3$\pm$2.2 & 83.1$\pm$1.9
& 88.5$\pm$0.9 & 1.7$\pm$1.5 & 90.4$\pm$0.8 \\

\rowcolor{babyblue}
\textsc{hidra} (Ours)
& \textbf{87.5$\pm$1.0} & -1.8$\pm$1.2 & \textbf{88.7$\pm$0.7}
& \textbf{90.5$\pm$0.6} & -3.6$\pm$1.2 & \textbf{92.8$\pm$0.5} \\

\midrule
\multicolumn{7}{c}{\textbf{LIBERO-SPATIAL}} \\
\midrule

\multirow{2}{*}{Method}
& \multicolumn{3}{c}{3-pass}
& \multicolumn{3}{c}{5-pass} \\
\cmidrule(lr){2-4}\cmidrule(lr){5-7}
& FWT$\uparrow$ & NBT$\downarrow$ & AUC$\uparrow$
& FWT$\uparrow$ & NBT$\downarrow$ & AUC$\uparrow$ \\
\midrule

MLR-IFA~\cite{yu2026lifelong}
& 59.0$\pm$2.4 & 8.8$\pm$1.3 & 52.9$\pm$1.2
& 62.0$\pm$1.8 & 9.0$\pm$2.1 & 57.6$\pm$2.0 \\

ISCIL~\cite{lee2024incremental}
& 56.0$\pm$2.8 & 1.7$\pm$3.3 & 54.6$\pm$2.5
& 54.0$\pm$2.7 & \textbf{-2.8$\pm$2.9} & 55.7$\pm$2.4 \\

DMPEL (100\%)~\cite{lei2025dynamic}
& \textbf{67.5$\pm$2.4} & 9.3$\pm$2.5 & 60.8$\pm$1.8
& 71.0$\pm$2.1 & 1.2$\pm$1.8 & 70.0$\pm$1.4 \\

\rowcolor{babyblue}
\textsc{hidra} (Ours)
& 62.0$\pm$2.7 & \textbf{-3.1$\pm$2.4} & \textbf{65.0$\pm$2.6}
& \textbf{74.5$\pm$2.2} & 4.2$\pm$3.5 & \textbf{72.5$\pm$2.3} \\

\midrule
\multicolumn{7}{c}{\textbf{LIBERO-GOAL}} \\
\midrule

\multirow{2}{*}{Method}
& \multicolumn{3}{c}{3-pass}
& \multicolumn{3}{c}{5-pass} \\
\cmidrule(lr){2-4}\cmidrule(lr){5-7}
& FWT$\uparrow$ & NBT$\downarrow$ & AUC$\uparrow$
& FWT$\uparrow$ & NBT$\downarrow$ & AUC$\uparrow$ \\
\midrule

MLR-IFA~\cite{yu2026lifelong}
& 76.0$\pm$2.9 & 1.8$\pm$1.9 & 74.3$\pm$2.0
& 77.0$\pm$1.8 & 10.0$\pm$4.4 & 70.6$\pm$2.8 \\

ISCIL~\cite{lee2024incremental}
& 59.5$\pm$2.4 & \textbf{-6.4$\pm$3.6} & 63.6$\pm$2.8
& 69.0$\pm$2.9 & 1.6$\pm$1.7 & 66.8$\pm$2.6 \\

DMPEL (100\%)~\cite{lei2025dynamic}
& 78.5$\pm$2.5 & -2.1$\pm$3.2 & 80.1$\pm$1.9
& 81.5$\pm$2.4 & \textbf{-0.1$\pm$3.6} & 80.7$\pm$3.1 \\

\rowcolor{babyblue}
\textsc{hidra} (Ours)
& \textbf{79.0$\pm$1.3} & -1.6$\pm$2.2 & \textbf{82.5$\pm$2.7}
& \textbf{84.5$\pm$1.3} & 1.1$\pm$2.6 & \textbf{83.6$\pm$2.1} \\

\midrule
\multicolumn{7}{c}{\textbf{LIBERO-10}} \\
\midrule

\multirow{2}{*}{Method}
& \multicolumn{3}{c}{3-pass}
& \multicolumn{3}{c}{5-pass} \\
\cmidrule(lr){2-4}\cmidrule(lr){5-7}
& FWT$\uparrow$ & NBT$\downarrow$ & AUC$\uparrow$
& FWT$\uparrow$ & NBT$\downarrow$ & AUC$\uparrow$ \\
\midrule

MLR-IFA~\cite{yu2026lifelong}
& 61.0$\pm$2.6 & 0.5$\pm$1.6 & 60.7$\pm$1.9
& 71.0$\pm$1.9 & 11.9$\pm$3.8 & 61.5$\pm$2.6 \\

ISCIL~\cite{lee2024incremental}
& 49.5$\pm$3.8 & 4.3$\pm$4.3 & 45.4$\pm$3.5
& 56.0$\pm$3.0 & 14.9$\pm$5.2 & 43.8$\pm$3.6 \\

DMPEL (100\%)~\cite{lei2025dynamic}
& 54.5$\pm$3.7 & 1.0$\pm$1.5 & 54.4$\pm$2.3
& 60.5 $\pm$ 2.0 & 0.2 $\pm$ 1.4 & 60.5 $\pm$ 2.7 \\

\rowcolor{babyblue}
\textsc{hidra} (Ours)
& \textbf{63.0$\pm$2.0} & \textbf{-4.0$\pm$3.2} & \textbf{64.9$\pm$2.9}
& \textbf{68.5$\pm$2.7} & \textbf{-1.0$\pm$3.4} & \textbf{69.1$\pm$3.2}
 \\

\bottomrule
\end{tabular}}
\end{table}

While our method targets a single-pass setting, we additionally evaluate it under multiple training passes to assess robustness beyond this regime. Specifically, each task is trained for 3--5 epochs, following prior work that allows limited data revisitation after large-scale pre-training (e.g., LIBERO-90). We compare against the strongest baselines from Tab.~\ref{tab:libero_main}.

As shown in Tab.~\ref{tab:epoch}, our method achieves the best AUC in most settings and consistently strong forward transfer. The gains are particularly evident in the 5-pass regime, where performance improves further across all benchmarks. While some baselines achieve lower NBT in specific cases, our method remains competitive and provides a better overall balance between adaptation and retention (i.e., higher overall AUC).

Overall, these results indicate that the proposed approach generalizes well beyond the single-pass regime, maintaining strong performance even when multiple passes over the data are allowed.

\subsection{Further Analysis Related to Routing Mechanism}
\begin{figure}[ht!]
    \centering
    \includegraphics[width=0.5\linewidth]{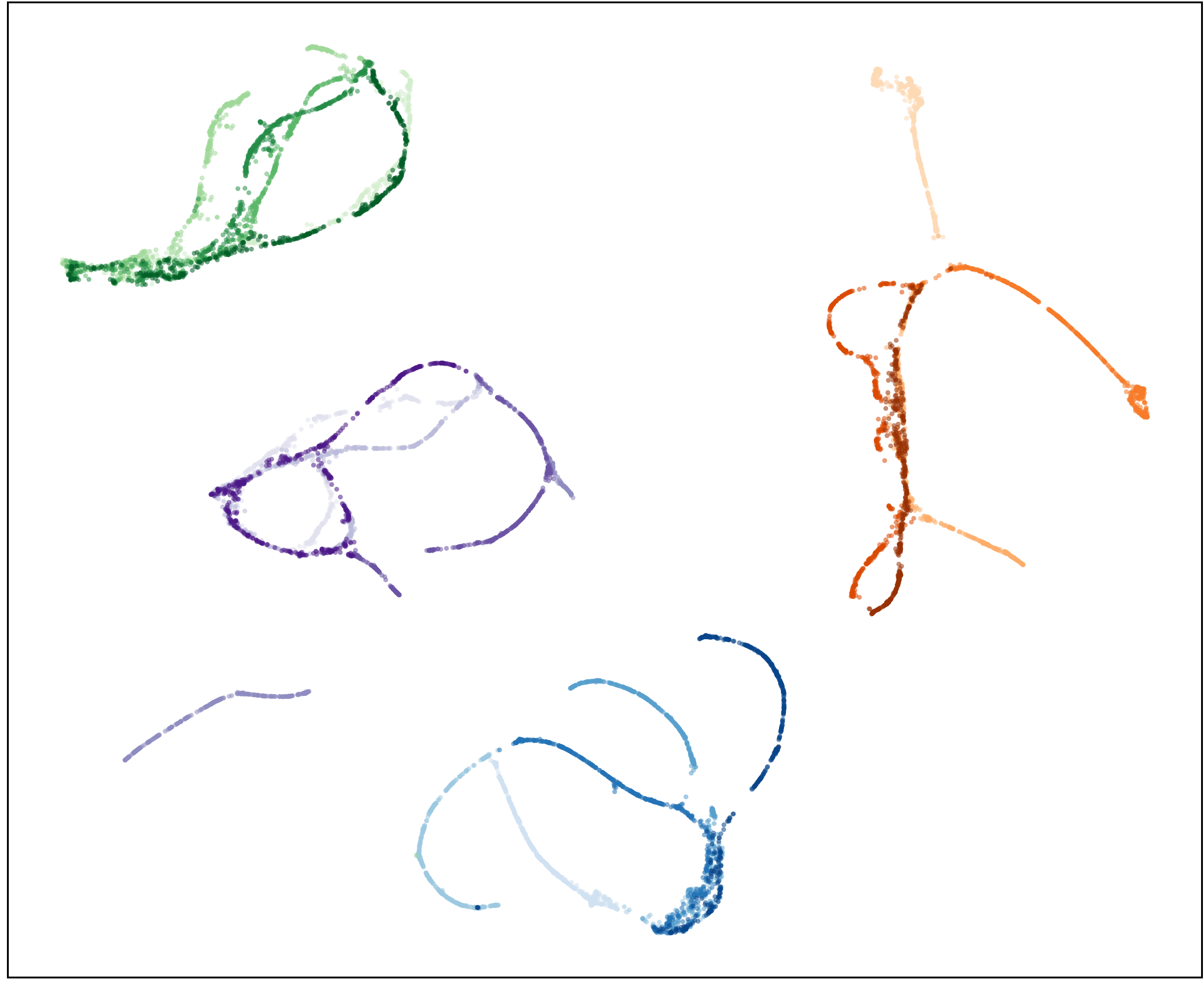}
    \caption{UMAP visualization of context query representations for all tasks in LIBERO-MIX.}
    \label{fig:query_vis}
\end{figure}

We visualize the context query representations of all tasks in LIBERO-MIX using UMAP in Fig.~\ref{fig:query_vis}. Each task is marked with a distinct color, while tasks from the same benchmark (i.e., OBJECT, SPATIAL, GOAL, LONG) are assigned colors from the same color family. For example, the tasks from LIBERO-GOAL are shown using different shades of green (e.g., light green, dark green). As observed in Fig.~\ref{fig:query_vis}, tasks with similar semantics tend to form nearby clusters in the embedding space.
 
While this structure is beneficial for capturing shared properties, it also introduces a potential issue: task representations may become too similar, leading to ambiguity during expert selection.
In particular, if the context keys of two tasks are highly similar, i.e., 
$k_i^{\mathrm{ctx}} \approx k_j^{\mathrm{ctx}}$, 
then for a given query $q_t^{\mathrm{ctx}}$, their attention scores will also be similar 
($a_{i,t}^{\mathrm{ctx}} \approx a_{j,t}^{\mathrm{ctx}}$). 
This makes it difficult for the routing mechanism to distinguish between the corresponding experts, causing them to be selected or ignored together.
Such behavior reduces the discriminability of expert selection and limits the diversity of expert combinations, 
which is particularly problematic in lifelong learning where tasks accumulate over time. 
To address this issue, our inter-task regularization, which explicitly encourages separation between context keys of different tasks while preserving their representational fidelity, is useful, as shown empirically in the ablation study.

\subsection{Task Instructions used in LIBERO-INS and LIBERO-MIX}
\label{app:task_instructions}

The full task lists used in LIBERO-INS and LIBERO-MIX are shown in Tab.~\ref{tab:liberoins} and Tab.~\ref{tab:libero_mix_tasks}, respectively, to facilitate reproducibility.

\begin{table*}[tb!]
\centering
\caption{Original and paraphrased instructions used for the LIBERO-INS.}
\label{tab:liberoins}
\small
\begin{tabularx}{\textwidth}{cXX}
\toprule
\textbf{Task} & \textbf{Original instruction during adaptation} & \textbf{Paraphrased instruction during testing} \\
\midrule
1 &
put both the alphabet soup and the tomato sauce in the basket &
place the alphabet soup and the tomato sauce into the basket \\
2 &
put both the cream cheese box and the butter in the basket &
move the cream cheese box and the butter into the basket \\
3 &
turn on the stove and put the moka pot on it &
switch on the stove and place the moka pot on it \\
4 &
put the black bowl in the bottom drawer of the cabinet and close it &
put the black bowl into the lower drawer of the cabinet, then close the drawer \\
5 &
put the white mug on the left plate and put the yellow and white mug on the right plate &
place the white mug on the left plate and the yellow and white mug on the right plate \\
6 &
pick up the book and place it in the back compartment of the caddy &
pick up the book and put it into the rear compartment of the caddy \\
7 &
put the white mug on the plate and put the chocolate pudding to the right of the plate &
place the white mug on the plate and put the chocolate pudding to the right side of the plate \\
8 &
put both the alphabet soup and the cream cheese box in the basket &
place both the alphabet soup and the cream cheese box into the basket \\
9 &
put both moka pots on the stove &
put the two moka pots on the stove \\
10 &
put the yellow and white mug in the microwave and close it &
put the yellow and white mug inside the microwave and close the microwave door \\
\bottomrule
\end{tabularx}
\end{table*}

\begin{table*}[tb!]
\centering
\caption{Tasks used in LIBERO-MIX.}
\label{tab:libero_mix_tasks}
\small
\begin{tabularx}{\textwidth}{cX}
\toprule
\textbf{Task} & \textbf{Instruction} \\
\midrule
1  & pick up the black bowl between the plate and the ramekin and place it on the plate \\
2  & pick up the alphabet soup and place it in the basket \\
3  & open the middle drawer of the cabinet \\
4  & put both the alphabet soup and the tomato sauce in the basket \\
5  & pick up the black bowl next to the ramekin and place it on the plate \\
6  & pick up the cream cheese and place it in the basket \\
7  & put the bowl on the stove \\
8  & put both the cream cheese box and the butter in the basket \\
9  & pick up the black bowl from table center and place it on the plate \\
10 & pick up the salad dressing and place it in the basket \\
11 & put the wine bottle on top of the cabinet \\
12 & put the white mug on the left plate and put the yellow and white mug on the right plate \\
13 & pick up the black bowl on the cookie box and place it on the plate \\
14 & pick up the bbq sauce and place it in the basket \\
15 & open the top drawer and put the bowl inside \\
16 & put the white mug on the plate and put the chocolate pudding to the right of the plate \\
17 & pick up the black bowl in the top drawer of the wooden cabinet and place it on the plate \\
18 & pick up the ketchup and place it in the basket \\
19 & put the bowl on top of the cabinet \\
20 & put both the alphabet soup and the cream cheese box in the basket \\
\bottomrule
\end{tabularx}
\end{table*}

\subsection{Statistical Significance}
\label{sec:statisticalSig}

To assess the robustness of our results, we perform pairwise comparisons using Welch’s t-test based on the reported mean and standard deviation across three seeds in Tab. \ref{tab:libero_main}. HIDRA shows statistically significant improvements in AUC over most baselines, particularly on LIBERO-OBJECT and LIBERO-SPATIAL, where performance gains are large. Compared to strong baselines such as DMPEL, improvements remain consistent across benchmarks, although differences are not always statistically significant in settings where performance gaps are small (e.g., LIBERO-GOAL and LIBERO-10). Overall, these results indicate that HIDRA provides reliable and consistent performance gains, with strongest significance observed in more challenging settings (Tab. \ref{tab:significance}).

\begin{table}[t]
\centering
\caption{Approximate statistical significance (Welch’s t-test) for AUC between HIDRA and representative baselines. $^*$ indicates $p < 0.05$.}
\label{tab:significance}
\begin{tabular}{lcccc}
\toprule
 \textsc{HIDRA} & OBJECT & GOAL & SPATIAL & LIBERO-10 \\
\midrule
vs MLR-IFA 
& $<0.001^*$ & $<0.001^*$ & $<0.001^*$ & $<0.001^*$ \\

vs ISCIL 
& $<0.001^*$ & $<0.001^*$ & $<0.001^*$ & $<0.001^*$ \\
vs DMPEL (100\%) 
& $<0.001^*$ & $0.18$ & $<0.001^*$ & $0.09$ \\

vs DMPEL (50\%) 
& $<0.001^*$ & $0.07$ & $<0.001^*$ & $0.06$ \\
\bottomrule
\end{tabular}
\end{table}

\subsection{Routing stability under library expansion}
Previously learned experts, routing projections, and their associated keys remain frozen, so the scores of an old input against existing experts do not change as the expert library expands. However, a newly introduced instruction key becomes an additional Stage-I competitor and may, in principle, replace the previously selected anchor. The anchor-conditioned predecessor set prevents later experts from entering Stage II as long as the original anchor remains selected, but it does not provide a formal guarantee against such anchor replacement. In our LIBERO-OBJECT analysis, we observed zero retroactive anchor switches as the expert library expanded. The fixed task instructions used by standard LIBERO and the frozen representation of previously learned inputs help make Stage-I retrieval stable, but we restrict this statement to empirical stability rather than claiming a formal guarantee.

Similarly, the newly created expert is not enforced to participate during its own adaptation through a hard inclusion constraint. When task $T_j$ arrives, however, the instruction key of the new expert is initialized from the normalized instruction representation of $T_j$, yielding an initial cosine similarity of $1$ with the corresponding instruction query and therefore selecting the expert at the beginning of adaptation. The task instruction remains fixed, while the instruction key is updated using $0.01\times$ the learning rate of the other trainable parameters. Empirically, the newly created expert remained selected throughout its own adaptation and thus received policy-loss gradients in all our experiments, although this behavior is not mathematically guaranteed for arbitrary task sequences.

\subsection{Limitations and Future Work}
\label{sec:limitFuture}

While HIDRA achieves strong performance in replay-free lifelong imitation learning, several aspects provide opportunities for further extension. As observed in the experiments, performance is lower on tasks requiring fine-grained spatial reasoning (e.g., LIBERO-SPATIAL), where expert selection relies more on context representations than the instruction-level. This reflects a broader challenge in multimodal robotic learning, where capturing subtle spatial relationships remains difficult for existing approaches; notably, competing methods exhibit similar or larger performance drops in these settings. Improving spatial representations and context modeling is, therefore a promising direction for future work.

Experiments are conducted on the LIBERO benchmark, which, although diverse and widely used, remains a simulated environment similar to prior work. At the same time, we extend standard evaluation by introducing more challenging settings such as LIBERO-MIX and LIBERO-INS, which incorporate heterogeneous task sequences and language variation not considered in prior studies. Evaluating the method on real robotic platforms would further assess its robustness under real-world conditions such as sensor noise and actuation variability. In addition, while the expandable expert design enables continual growth, scaling to very large numbers of tasks (e.g., millions) may require more efficient routing or expert management strategies, a challenge shared by other modular and expert-based methods such as TAIL and DMPEL.

Overall, these considerations highlight natural directions for extending the proposed framework while reinforcing its effectiveness within the current experimental scope.

\end{document}